%% file: main.tex
\documentclass[lettersize,journal]{IEEEtran}

\ifCLASSOPTIONcompsoc
  \usepackage[nocompress]{cite}
\else
  \usepackage{cite}
\fi

\ifCLASSINFOpdf
\else
\fi

\usepackage{amsmath,amsfonts}
\usepackage{algorithmic}
\usepackage{algorithm}
\usepackage{array}
\usepackage[caption=false,font=normalsize,labelfont=sf,textfont=sf]{subfig}
\usepackage{textcomp}
\usepackage{stfloats}
\usepackage{url}
\usepackage{verbatim}
\usepackage{graphicx}
\usepackage{cite}
\usepackage[table,x11names]{xcolor}
\usepackage{multirow}
\usepackage[colorlinks,linkcolor=red]{hyperref}
\newcommand{\ie}{\textit{i.e.}}

\newcommand{\xieming}[1]{\textcolor{black}{#1}} 

\begin{document}

\title{AnyRefill: A Unified, Data-Efficient Framework for Left-Prompt-Guided  Vision Tasks}

\author{
{Ming Xie$^{\star}$, Chenjie Cao$^{\star}$, Yunuo Cai, Xiangyang Xue, Yu-Gang Jiang, Yanwei Fu$^{\dag}$}
\thanks{$^\star$ indicates equal contributions, $^\dag$ refers to the corresponding author.}
\IEEEcompsocitemizethanks{\IEEEcompsocthanksitem All the authors are affiliated with Fudan University, China.
Ming Xie and Yanwei Fu are also with the Shanghai Innovation Institute. 
Chenjie Cao is also with DAMO Academy, Alibaba Group. Email: \{mxie20,yncai20, xyxue, ygj, yanweifu\}@fudan.edu.cn,
 caochenjie.ccj@alibaba-inc.com.
}
}

\markboth{Journal of \LaTeX\ Class Files,~Vol.~14, No.~8, August~2021}%
{Shell \MakeLowercase{\textit{et al.}}: A Sample Article Using IEEEtran.cls for IEEE Journals}


\maketitle

\begin{abstract}
In this paper, we present a novel Left-Prompt-Guided (LPG) paradigm to address a diverse range of reference-based vision tasks. Inspired by the human creative process, we reformulate these tasks using a left-right stitching formulation to construct contextual input. Building upon this foundation, we propose AnyRefill, an extension of LeftRefill~\cite{cao2024leftrefill}, that effectively adapts Text-to-Image (T2I) models to various vision tasks.
AnyRefill leverages the inpainting priors of advanced T2I model based on the Diffusion Transformer (DiT) architecture, and incorporates flexible components to enhance its capabilities. By combining task-specific LoRAs with the stitching input, AnyRefill unlocks its potential across diverse tasks, including conditional generation, visual perception, and image editing, without requiring additional visual encoders.
Meanwhile, AnyRefill exhibits remarkable data efficiency, requiring minimal task-specific fine-tuning while maintaining high generative performance.
Through extensive ablation studies, we demonstrate that AnyRefill outperforms other image condition injection methods and achieves competitive results compared to state-of-the-art open-source methods.
Notably, AnyRefill delivers results comparable to advanced commercial tools, such as IC-Light and SeedEdit as shown in  Fig.~\ref{fig:gender_editing_compare},~\ref{fig:editing_relight}, even in challenging scenarios. Comprehensive experiments and ablation studies across versatile tasks validate the strong generation of the proposed simple yet effective LPG formulation, establishing AnyRefill as a unified, highly data-efficient solution for reference-based vision tasks.
\end{abstract}


\section{Introduction}
\begin{figure}[h!]
    \centering
    \includegraphics[width=0.42\textwidth]{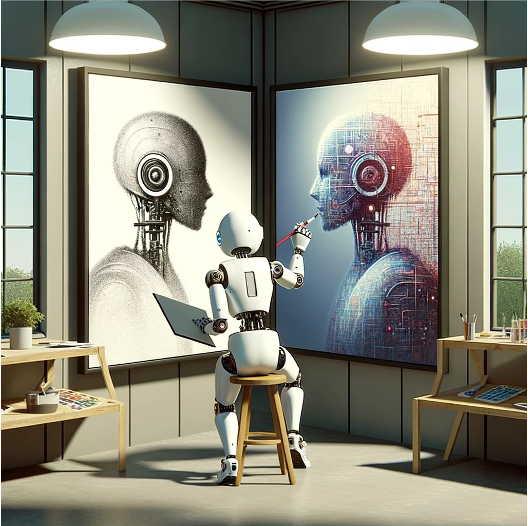} 
    \vspace{-0.05in}
    \caption{An image, generated by DALL·E 3~\cite{betker2023improving}, vividly illustrates the motivation behind AnyRefill. A robot, representing the T2I model in AnyRefill, acts as an experienced painter, using the left image as a reference to create content on the right canvas.}
    \vspace{-0.1in}
    \label{fig:teaser_robot}
\end{figure}

\begin{figure*}[!t]
    \centering
    \includegraphics[width=0.96\textwidth]{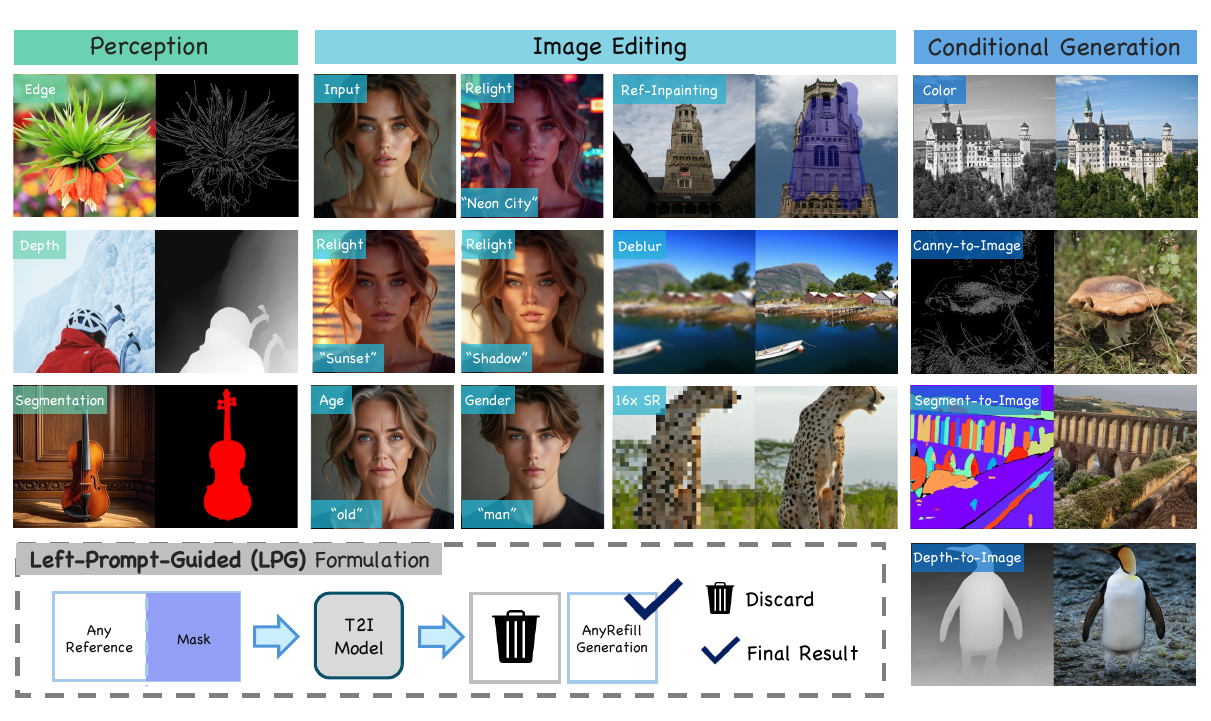} 
    \vspace{-0.15in}
    \caption{AnyRefill unifies various vision tasks by generating the right canvas conditioned by left references. We can re-formulate several existing tasks in the \textbf{Left-Prompt-Guided (LPG)} manner, including \textcolor[rgb]{0.42,0.82,0.70}{\textbf{perception}} tasks, \textcolor[rgb]{0.52,0.84,0.9}{\textbf{image editing}} tasks, and \textcolor[rgb]{0.39,0.65,0.89}{\textbf{conditional generation}} tasks.}
    \vspace{-0.1in}
    \label{fig:teaser}
\end{figure*}

\begin{table*}[h]
\small 
\caption{\underline{\textbf{Up}}: Tasks tackled by AnyRefill using few-shot data pairs, highlighting AnyRefill is a data-efficient method. ``Min Train Pairs'' means the minimum pairs used in AnyRefill experiments to explore the lower bound of data requirements to achieve reliable results. \underline{\textbf{Down}}: Tasks tackled by AnyRefill using sufficient data pairs, denoting that AnyRefill enjoys ability to be scaled up for large-scale training. 
\label{tab:teaser_table} }
\centering
\begin{tabular}{c|c|cccc}
     \hline
     \multicolumn{2}{c|}{Tasks} & {Min Train Pairs} & {Train Steps} & {Visualization} \\
     \hline
     \multirow{4}{*}{Generation}
     & Canny-to-Image & 10 & 10k  & Fig.~\ref{fig:generation_task_results} \\
     & Depth-to-Image & 10 & 10k  & Fig.~\ref{fig:generation_task_results} \\
     & Segment-to-Image & 10 & 10k  & Fig.~\ref{fig:generation_task_results} \\
     & Colorization & 10 & 20k  & Fig.~\ref{fig:colorize_editing} \\
     \hline
     \multirow{4}{*}{Image Editing}
     & Deblur & 100 & 10k  & Fig.~\ref{fig:generation_deblurring} \\
     & Age Editing & 50 & 10k  & Fig.~\ref{fig:gender_editing_compare} \& Fig.~\ref{fig:multi_lora_editing} \\
     & Gender Editing & 50 & 8k  & Fig.~\ref{fig:gender_editing_compare} \& Fig.~\ref{fig:multi_lora_editing} \\
     & Relighting & 35 & 8k & Fig.~\ref{fig:editing_relight} \\
     \hline
\end{tabular}
\renewcommand\tabcolsep{1.75pt}
\\ [0.35cm]
\begin{tabular}{c|c|cccc}
     \hline
     \multicolumn{2}{c|}{Tasks} & Train Pairs & Train Steps & Metric & Visualization \\
     \hline
     \multirow{3}{*}{Perception}
     & Image-to-Canny & 3450 & 15k & - & Fig.~\ref{fig:generation_perception} \\
     & Image-to-Depth & 3450 & 15k & - & Fig.~\ref{fig:generation_perception} \\
     & Image-to-Segment & 6125 & 15k & - & Fig.~\ref{fig:generation_perception} \\
     \hline
     \multirow{2}{*}{Image Editing}
     & Ref-Inpainting & 820k & 90k & Tab.~\ref{tab:reference_inpainting_quantitative} & Fig.~\ref{fig:reference_guided_qualitative} \\
     & Super Resolution (4x, 8x, 16x) & 3450 & 10k & Tab.~\ref{tab:quantitative_super_resolution} & Fig.~\ref{fig:generation_super_resolution} \\
     \hline
\end{tabular}
\end{table*}

\IEEEPARstart{I}{magine} being a right-handed painter tasked with creating or modifying a target image based on a reference picture. Naturally, you would place the reference image on your left side for easy access and use it as a guide while working on the right canvas\footnote{Of course, the entire process can be mirrored by swapping left and right for left-handed users.}. This intuitive spatial arrangement reflects how visual relationships are often structured in guided vision tasks.
Based on this idea, we introduce a novel and unified \textbf{Left-Prompt-Guided (LPG)} paradigm for reference-based vision tasks. Our framework leverages large visual foundation models, particularly Text-to-Image (T2I) models, where the left-side image serves as a visual prompt to guide contextual inpainting or synthesis on the right-side canvas, as illustrated in the lower-left corner of Fig.~\ref{fig:teaser}. This paradigm allows us to design a versatile, data-efficient model—dubbed \textit{AnyRefill}—that can effectively tackle a wide range of vision tasks within the LPG framework.
Given the impressive capabilities of state-of-the-art T2I models~\cite{nichol2021glide,rombach2022high,ramesh2022hierarchical,saharia2022photorealistic,chang2023muse,podell2023sdxl}, which act as skillful digital painters, an important question arises: Can these models be adapted to follow the intuitive LPG formulation and handle complex, reference-guided vision tasks with data efficiency?

It seems straightforward to harness the power of T2I generative models to directly address these reference-based vision tasks by training additional adapters~\cite{hu2021lora,zhang2023adding,mou2023t2i} or replacing textual encoders with visual ones~\cite{yang2023paint,liu2023zero}
and optimize them for full fine-tuning of the entire T2I model.
We should clarify that training these large T2I models with `unfamiliar' visual encoders is computationally intensive and challenging to converge, particularly when working with limited batch sizes. Additionally, most visual encoders, such as image CLIP~\cite{radford2021learning}, are primarily designed to capture high-level semantic features rather than the intricately spatial details that are essential for tasks involving Ref-inpainting. This limitation underscores the need for more efficient frameworks that are specifically attentive to spatial conditioning for synthesis tasks.

To avoid extensive modifications in Text-to-Image (T2I) models, we introduce \textit{AnyRefill}, a unified framework inspired by human painting intuition for LPG synthesis. Building on the prototype LeftRefill~\cite{cao2024leftrefill}, originally introduced in our earlier conference work, AnyRefill leverages advanced Diffusion Transformer (DiT) architecture~\cite{peebles2023scalable} and FLUX.Fill~\cite{fluxfill2024} to extend the LPG concept beyond the U-Net-based StableDiffusion (SD) inpainting~\cite{rombach2022high}\footnote{\scriptsize\url{https://github.com/Stability-AI/stablediffusion}}, significantly broadening its applicability.
AnyRefill reformulates reference-based synthesis as an LPG formulation inpainting or synthesis process, enabling it to effectively treat diverse vision tasks in an ``all-in-one'' manner, as shown in Fig.~\ref{fig:teaser} and listed below:
\begin{itemize}
    \item  \textbf{Conditional generation tasks:} canny-to-image generation, depth-to-image generation, segmentation-to-image generation~\cite{zhang2023adding}, as well as coloring~\cite{tan2024ominicontrol};
    \item \textbf{Perception tasks:} depth estimation~\cite{ke2023repurposing}, canny edge extraction, and foreground segmentation~\cite{zhao2023unleashing}),
    \item \textbf{Image editing tasks:} Ref-inpainting~\cite{zhou2021transfill,zhao2022geofill}, deblurring~\cite{tan2024ominicontrol}, super-resolution~\cite{wang2024exploiting,tan2024ominicontrol}, relighting~\cite{iclight2024}, and attribute modifications like gender editing and age editing~\cite{rout2024semantic, dalva2024fluxspace}).
\end{itemize}
All these tasks are summarized in Tab.~\ref{tab:teaser_table}. Critically, conditional generation tasks create new content from rough conditions, perception tasks extract perceptual information for image and scene understanding, and image editing tasks modify existing content to enhance image quality or adjust object attributes. 

The key innovation of AnyRefill, similar to LeftRefill, lies in its LPG formulation, where reference and target views are horizontally concatenated into a single input: reference images occupy the left side, while masked target regions are positioned on the right (Fig.~\ref{fig:teaser}). This streamlined design eliminates the need for additional image feature encoders or external meta-knowledge by integrating both views into a unified canvas. To enable AnyRefill to fully leverage the inpainting priors of T2I models and act as a professional painter, we equipped it with a task-specific LoRA~\cite{hu2021lora} to efficient fine-tune without compromising generative performance, allowing AnyRefill to reliably learn the LPG paradigm.

While high-quality data from commercial models is challenging to produce in bulk, AnyRefill integrates task-specific LoRA within our LPG formulation, and surprisingly exhibits exceptional data efficiency in reference-based vision tasks. 
We summarize the qualitative results of AnyRefill under few-shot scenarios in Tab.~\ref{tab:teaser_table} (Up) while Tab.~\ref{tab:teaser_table} (Down) shows results with sufficient data pairs, denoting that AnyRefill enjoys the ability to be scaled up for large-scale training. 
Notably, by leveraging pseudo-image pairs generated from specialized models, AnyRefill not only surpasses publicly available methods~\cite{wang2018esrgan} but also delivers results comparable to advanced commercial tools~\cite{iclight2024}, including the state-of-the-art image editing model SeedEdit~\cite{shi2024seededit}.
Despite the distinct goals of these tasks shown in Tab.~\ref{tab:teaser_table}, AnyRefill achieves remarkable adaptability in all reference-based vision tasks using limited training pairs, which we attribute to the contextual richness provided by the LPG paradigm and inpainting priors. 
In addition, our ablation studies in Sec.~\ref{sec:ablation_controlnet_ipadapter} demonstrate that the LPG formulation outperforms other widespread image condition
injection approaches~\cite{zhang2023adding,mou2023t2i} when training data is limited. AnyRefill utilizes task-specific low-rank matrices to inject crucial guidance into the attention modules of the DiT model, steering the generative process.

Another significant strength of AnyRefill is its flexibility and efficiency. Unlike traditional approaches that require specialized model architecture for individual tasks~\cite{peebles2023scalable,goodfellow2020generative,dosovitskiy2020image,he2016deep,ronneberger2015u}, AnyRefill unifies these tasks within a single framework by employing general LPG formulation. This generalization endows AnyRefill with greater potential for practical capability.

Our contributions can be summarized as follows:

\noindent(1) \textbf{Proposing LPG Paradigm as a Unified and Simplified Design}: Inspired by human painting, the Left-Prompt-Guided (LPG) paradigm structures reference-based vision tasks with the reference image on the left and the target on the right. This simple yet effective approach unifies contextual inpainting and synthesis tasks within a single framework. By horizontally stitching input views, LPG eliminates the need for extra encoders or external knowledge, enabling independent training for diverse tasks without test-time fine-tuning

\noindent(2) \textbf{Presenting the AnyRefill Framework: Unifying Vision Tasks with a Single Model}: 
Building upon LeftRefill~\cite{cao2024leftrefill} and the LPG paradigm, AnyRefill is a data-efficient image generation model that leverages T2I models and incorporates advanced DiT-based FLUX.Fill as component. AnyRefill uniquely addresses diverse vision tasks, including Conditional Generation, Perception, and Image Editing, within a single unified framework.

\noindent(3)\textbf{ High-Quality Results with Efficiency and Scalability}: 
AnyRefill exhibits remarkable data efficiency, requiring minimal task-specific fine-tuning while maintaining high generative performance. It excels in few-shot scenarios and scales effectively with larger datasets. Outperforming existing methods, AnyRefill achieves results comparable to advanced commercial tools. Its contextual richness and efficient inpainting priors enable adaptability and high performance across diverse tasks. 

These contributions collectively establish AnyRefill as a unified, efficient, and scalable solution for reference-guided vision tasks.

\section{Related Work}
\subsection{Text-to-Image Generation and Controllability}
Diffusion model~\cite{sohl2015deep,ho2020denoising} has emerged as a foundational approach in generation tasks, particularly excelling in T2I synthesis. LDM~\cite{rombach2022high} further optimizes the process by operating in a compressed latent space rather than directly on high-dimensional pixel space, significantly improving computational efficiency and image fidelity. 
Moreover, DiT~\cite{peebles2023scalable} introduces a transformer-based architecture for diffusion processes, enabling enhanced scalability and flexibility. 
Recent achievements, such as FLUX~\cite{flux2024} and SD3~\cite{esser2024scaling}, further incorporate Multimodal-DiT (MM-DiT) and rectified flow sampling~\cite{liu2022flow} to achieve state-of-the-art performance. 

In parallel, autoregressive models have gained prominence in T2I too, applying techniques like VQ-VAE~\cite{van2017neural} and VQ-GAN~\cite{esser2021taming} to quantize images into discrete token sequences for language-like processing. 
Furthermore, visual autoregressive (VAR)~\cite{tian2024visual} forms a new paradigm to accomplish next-scale prediction, achieving fine-grained text-to-image alignment. 
However, these models could only be controlled by natural languages. As ``an image is worth hundreds of words'', T2I models based on natural texts fail to produce images with specific textures, locations, identities, and appearances~\cite{gal2022image}. 

Many works focus on image-guided generation~\cite{voynov2022sketch,li2023gligen,ma2023unified}. DreamBooth~\cite{ruiz2022dreambooth} personalizes T2I models by fine-tuning the whole model on custom data for specific objects or styles adaptation. ControlNet~\cite{zhang2023adding} and T2I-Adapter~\cite{mou2023t2i} learn trainable adapters~\cite{houlsby2019parameter} to inject visual clues to pre-trained T2I models without losing generalization and diversity.
But these moderate methods only work for simple style transfers. More spatially complex tasks, such as Ref-inpainting, are difficult to handle by ControlNet as verified in Sec.~\ref{sec:experiments}. 
Compared with these aforementioned manners, AnyRefill and its precursor, LeftRefill, enjoy spatial modeling capability simply by modifying the input, without requiring complex mechanisms.

\subsection{Parameter-Efficient Fine-tuning (PEFT)}
With the development of T2I models' capacities~\cite{podell2024sdxl,esser2024scaling,flux2024}, fine-tuning them for personal requirements is intolerable. Thus PEFT is proposed to address this issue with minimal computational overhead.

Textual inversion~\cite{gal2022image,mokady2022null} is an advanced technique for customized content generation, focusing on learning textual embeddings to represent new concepts. Prompt Tuning ~\cite{lester2021power,liu2021gpt,liu2021p} indicates fine-tuning token embeddings for transformers with frozen backbone to preserve the capacity. Prompt tuning is first explored for adaptively learning suitable prompt features for language models rather than manually selecting them for different downstream tasks~\cite{liu2023pre}. Moreover, prompt tuning has been further investigated in vision-language models~\cite{radford2021learning,ge2022domain} and discriminative vision models~\cite{jia2022visual,liao2023rethinking}.
Visual prompt tuning in~\cite{sohn2022visual} prepends trainable tokens before the visual sequence for transferred generations. 
Though both LeftRefill and~\cite{sohn2022visual} aim to tackle image synthesis, our prompt tuning is used for controlling text encoders rather than visual ones. 
Thus LeftRefill enjoys more intuitive prompt initialization from task-related textual descriptions.

LoRA~\cite{hu2021lora} is also a PEFT method that introduces additional low-rank matrices to certain linear layers of the model, which adjusts output distribution towards target tasks. RealFill~\cite{tang2024realfill} tackles image completion through test-time optimization at the instance level, adopting DreamBooth’s reconstruction process and incorporating learnable LoRA to avoid fine-tuning the entire model. By training on a few multi-view images for each inference time, it inpaints specific target views. In contrast, AnyRefill focuses on task-specific optimization at the task level, leveraging inpainting priors combined with stitching input and training LoRA to adapt T2I models to a variety of vision tasks with limited training data.

\subsection{Reference-guided Image Generation}
Image inpainting is a long-standing vision generation task, which aims to fill missing image regions with coherent results. Significant advancements have been made by both traditional approaches~\cite{bertalmio2000image,criminisi2003object,hays2007scene} and learning-based methods~\cite{zeng2020high,zhao2021large,li2022mat,suvorov2022resolution,dong2022incremental}.
Furthermore, Ref-inpainting requires recovering a target image with one or several reference views from different viewpoints~\cite{tang2024realfill,oh2019onion}, which is useful for repairing old buildings or removing occlusions in popular attractions. However, Ref-inpainting often involves a complex, multi-step pipeline~\cite{zhou2021transfill,zhao2022geofill,zhao20223dfill}, including depth estimation, pose estimation, homography warping, and single-view inpainting.
The reliability of these pipelines is compromised when large missing regions result in inaccurate geometric pose estimations, which significantly degrade performance. Thus an end-to-end Ref-inpainting pipeline is highly desirable. This highlights the need for more streamlined, scalable, and resource-efficient reference-guided generation methods—a challenge effectively tackled by our proposed LPG framework.

\subsection{Image-to-Image Editing}
Image editing aims to modify specific content in an image based on text while preserving other regions unchanged. Training-free image editing methods have garnered increasing attention due to their convenience and efficiency. SDEdit~\cite{meng2021sdedit} innovatively adds noise to image up a specified step and denoises conditioning on a target prompt to get desired edit. Other training-free methods explore attention manipulation~\cite{hertz2022prompt,mokady2023null,parmar2023zero,dalva2024fluxspace}, mask guidance~\cite{avrahami2022blended,couairon2022diffedit,huang2023region,li2024zone}, or modifications to RF sampling processes~\cite{rout2024semantic,wang2024taming,kulikov2024flowedit}. Despite their advantages, the generative performance of training-free editing methods still lag behind supervised models~\cite{brooks2023instructpix2pix,sheynin2024emu,zhang2024hive,zhang2024magicbrush,shi2024seededit}. Supervised editing models require large and diverse image pairs for training, whereas AnyRefill strikes a balance between supervised and tuning-free approaches. By leveraging T2I inpainting priors, AnyRefill achieves competitive results with only a small amount of training data.

\subsection{Preliminaries of FLUX}
\label{sec:preliminary_anyrefill}

As our AnyRefill is built upon the FLUX model~\cite{flux2024}, we discuss the preliminaries of FLUX in this section.

\noindent\textbf{Rectified Flow (RF)~\cite{liu2022flow}.}
Generative models seek to learn a mapping from a noise distribution \( p_1 \) to a data distribution \( p_0 \), where \( p_0 \) typically represents real-world data such as images or videos, and \( p_1 \) is commonly chosen as a standard Gaussian distribution. RF proposes a simple yet effective approach to bridge these two distributions by constructing a direct trajectory in the latent space. This is accomplished by modeling a time-dependent flow governed by an Ordinary Differential Equation (ODE). Through simple linear interpolation, RF enables the velocity field to learn the process of gradually transitioning from real data distribution to noise one. Thus, in the inference time, the velocity field can iteratively generate real data distribution samples from noise distribution.

\noindent\textbf{Multimodal Diffusion Transformer (MM-DiT)~\cite{esser2024scaling}}
represents a notable advancement in multimodal generative models by effectively integrating both text and image modalities for text-guided image generation. 
Building upon the DiT framework~\cite{peebles2023scalable}, MM-DiT introduces two specialized mechanisms that facilitate robust multimodal interactions and ensure precise alignment between textual and visual content within a bidirectional flow: (1) \textit{SingleStream} block employs a unified attention mechanism to process concatenated text and image embeddings, capturing fine-grained semantic correlations. (2) \textit{DoubleStream} block separates text and image processing to preserve modality-specific information while enabling cross-modal interactions through shared intermediate layers.

As one of the leading T2I generation models, FLUX demonstrates exceptional text-image alignment capabilities by leveraging the advanced MM-DiT architecture. Furthermore, FLUX integrates textual embeddings from both CLIP-L~\cite{radford2021learning} and T5~\cite{raffel2020exploring}, ensuring the retention of rich textual semantics.

\noindent\textbf{FLUX.Fill~\cite{fluxfill2024}.}
Building on FLUX~\cite{flux2024}, FLUX.Fill is fine-tuned using additional masked latents and mask maps to address the inpainting task. Leveraging the powerful MM-DiT architecture, a larger model capacity (12B vs. 0.8B), and more extensive training data, FLUX.Fill delivers superior performance across all metrics compared to SD~\cite{rombach2022high}.
Inspired by prior research emphasizing the role of textual semantics in enhancing MM-DiT's generation quality~\cite{esser2024scaling,ramesh2021zero,saharia2022photorealistic}, we fine-tune FLUX.Fill following the LeftRefill paradigm, adopting LoRA~\cite{hu2021lora} rather than prompt tuning to preserve robust textual alignment capabilities.

\section{AnyRefill under Left-Prompt-Guided Formulation}

\noindent\textbf{Roadmap.} 
In this section, we first define and motivate LPG in Sec.~\ref{sec:lpg}. We then briefly review LeftRefill in Sec.~\ref{sec:prompt_tuning}, based on Diffusion U-Net. Next, we extend LPG to AnyRefill using the rectified flow-based DiT framework, FLUX~\cite{flux2024}, and provide the overview of different vision tasks in Sec.~\ref{section:task-specific}. 
Subsequently, details on task-specific fine-tuning and dataset construction are discussed in Sec.~\ref{sec:detail_anyrefill}.

\begin{figure}
\centering
\includegraphics[width=1.0\linewidth]{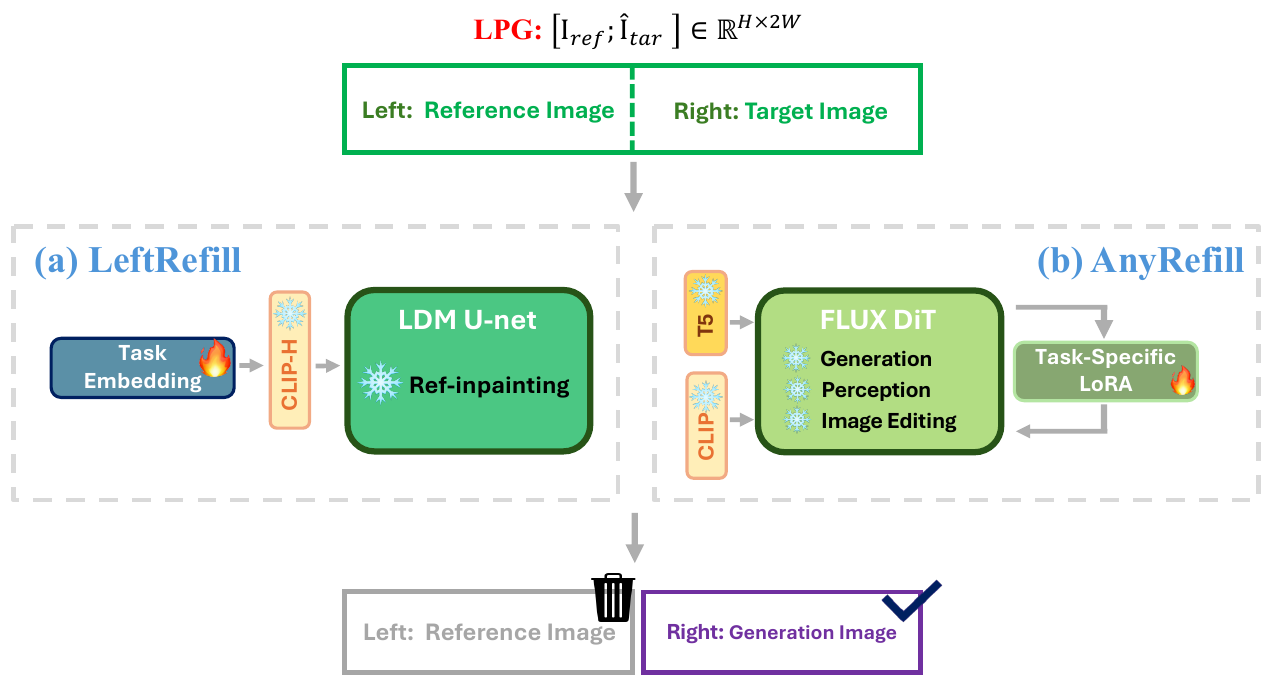}
\vspace{-0.2in}
   \caption{Overview and comparison of (a) our previous LeftRefill~\cite{cao2024leftrefill} and (b) AnyRefill under the LPG formulation. 
   The task prompt embedding is infused into CLIP-H for cross-attention learning in U-net, while task-specific LoRAs are adopted to the rectified flow-based DiT for more diverse vision tasks.
   For the output of LeftRefill and AnyRefill, we discard the left-side reference and take the right-side generation.
   \label{fig:model_overview}}
\vspace{-0.1in}
\end{figure}

\begin{figure}
\centering
\includegraphics[width=1.0\linewidth]{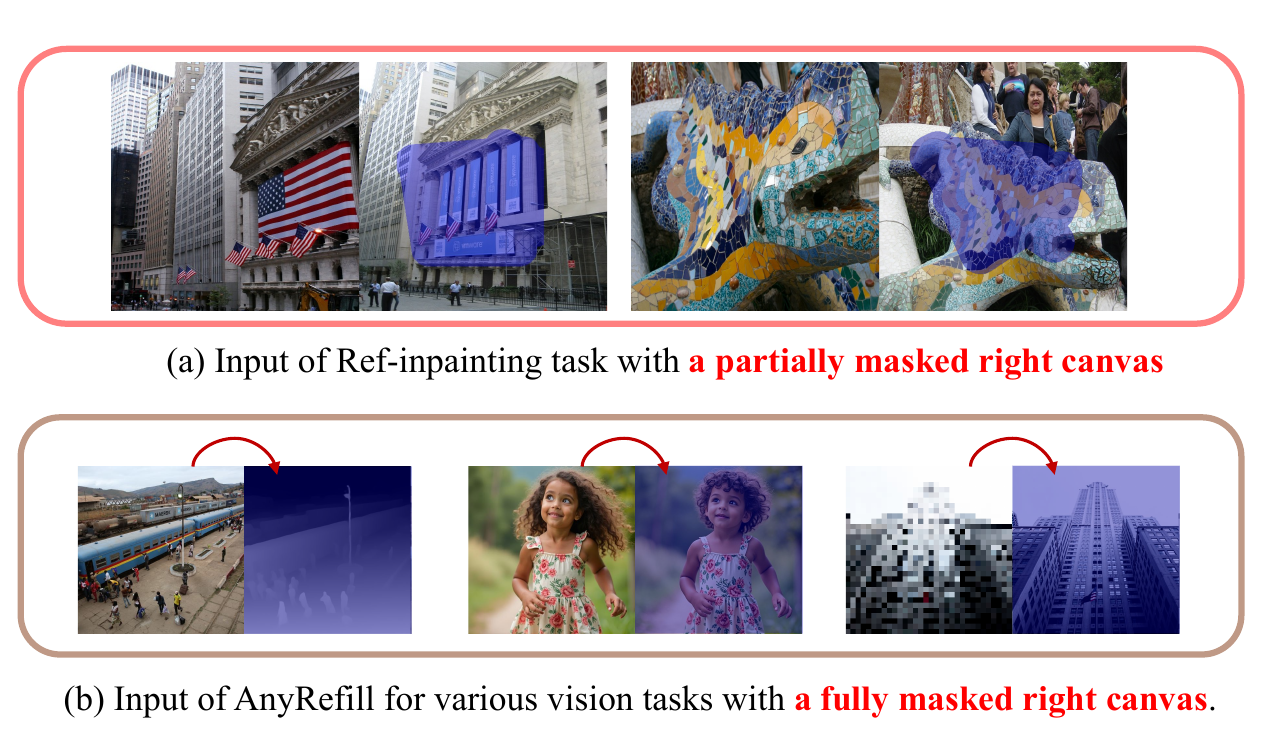}
\vspace{-0.2in}
   \caption{Inputs of (a) \textbf{Partially masked targets} for reference inpainting based tasks; and (b) \textbf{Fully masked right canvas} for various other vision tasks.
   Masked regions are indicated with \textcolor{blue!65}{semi-transparent blue}.
   \label{fig:input_paradigm}}
\vspace{-0.15in}
\end{figure}

\subsection{Left-Prompt-Guided (LPG) Formulation}
\label{sec:lpg}

\noindent\textbf{Definition of LPG.}
The overview of the proposed method under Left-Prompt-Guided formulation is depicted in Fig.~\ref{fig:model_overview}. 
In particular, the input image $\mathbf{I}'$ is formed by horizontally stitching the reference image $\mathbf{I}_{ref}$ and the masked target image $\hat{\mathbf{I}}_{tar}$ along the spatial dimension, expressed as $\mathbf{I}' = [\mathbf{I}_{ref}; \hat{\mathbf{I}}_{tar}] \in \mathbb{R}^{H \times 2W}$, as shown in the upper of Fig.~\ref{fig:model_overview}.
By default, the reference image, serves as a visual prompt, is positioned on the left, while the target image is placed on the right. 
The output image from the right half serves as the final generated result, while the left half is directly discarded.
Note that the diffusion optimization is based on the whole stitched image without any modification.
Besides, The masked target $\hat{\mathbf{I}}_{tar}$ is defined such that Ref-inpainting inputs are considered \emph{partially masked}, while the target images for other vision tasks are treated as \emph{entirely masked}, as shown in Fig.~\ref{fig:input_paradigm}.
Furthermore, the RoPE mechanism~\cite{su2024roformer} is expanded spatially within FLUX instead of interpolating the positional encoding map. 
It is a flexible framework that seamlessly switches between U-Net-based and DiT-based networks. The LPG-based LeftRefill employs task prompt embeddings for  parameter-efficient fine-tuning, while task-specific LoRAs are adopted in AnyRefill to handle more diverse vision tasks.
Consequently, the proposed LPG module serves as a key component in AnyRefill for constructing left-right stitched inputs for T2I models.

\noindent\textbf{Key Insights}. 
Two primary motivations make us stitch reference and target images together and reformulate diverse vision tasks as an LPG contextual inpainting problem.
First, AnyRefill operates with a single input image, thereby eliminating the need for additional image encoders, avoiding significant architectural modifications, and reducing the necessity for extensive re-training.
Second, since all T2I models are only pre-trained on single-view images, the left-right stitched input fomulation effectively reactivates the \emph{intrinsic capability of large T2I models to capture correlations within single-view images}.
Particularly, the LPG input structure facilitates self-attention modules in correctly attending to relevant regions from the left-side reference image during the initial stages of sampling process, as illustrated in Fig.~\ref{fig:attention_map_vis}.
Comprehensive evaluations of various reference-guided approaches, including SD and FLUX, are presented in Sec.~\ref{sec:exp_image_editing}.
Both LeftRefill and AnyRefill substantially outperform other competitors with high efficiency and fewer trainable parameters.
While AnyRefill benefits from a more advanced backbone architecture~\cite{flux2024}, the fundamental concept of LPG inpainting remains consistent with that of LeftRefill.

\subsection{Warming up: LeftRefill with Task Prompt Tuning}
\label{sec:prompt_tuning}

To provide deeper insights into the superior generalization performance of AnyRefill across a broader range of vision tasks compared to LeftRefill~\cite{cao2024leftrefill}, we first start our discussion with the fine-tuning strategy utilized by our LeftRefill. Specifically, LeftRefill employs learnable prompt embeddings as the textual component within the CLIP-H~\cite{radford2021learning} of Stable Diffusion, being applied to cross-attention blocks as shown in Fig.~\ref{fig:model_overview}(a).

Specifically, LeftRefill prepares a set of trainable text embeddings for different generative tasks. Though there are only a few trainable parameters (50 tokens of 0.05M trainable parameters), LeftRefill astonishingly finds that prompt tuning is sufficient to drive complex generative tasks such as Ref-inpainting, even with a frozen LDM backbone. The trainable task prompt embeddings $p_t$ are initialized as the averaged embedding of the natural task description. The optimization target is formulated as:
\begin{equation}\small
\negthickspace\{p_t\}_{*}=\mathop{\arg\min}\limits_{\{p_t\}}\mathbb{E}\left[\left\|\varepsilon-\varepsilon_\theta\left([z_t;\hat{z}_0;\mathbf{M}],c_{\phi}(p_t),t\right)\right\|^2\right]
\label{eq:pt_objective}
\end{equation}
where $\varepsilon_{\theta}(\cdot)$ denotes the noise estimated by LDM; $c_{\phi}(\cdot)$ represents the frozen CLIP-H, $z_t$ is the noisy latent feature at step $t$ derived from input $z_0$, and $\hat{z}_0 = z_0 \odot (1-\mathbf{M})$ denotes the masked latent features concatenated with $z_t$ using the mask $\mathbf{M}$. This approach offers both training efficiency and parameter savings~\cite{lester2021power}.

Although prompt tuning is effective in U-Net-based LeftRefill with sufficient training data, we clarify that it would hinder the generalization for the MM-DiT~\cite{esser2024scaling} based AnyRefill with restricted training data. Detailed ablation studies are discussed in Sec.~\ref{sec:ablation}.

\subsection{Overview of Tasks Addressed by AnyRefill}
\label{section:task-specific}
To verify the effectiveness of the LeftRefill and extend the LPG concept to AnyRefill, which takes left-right stitching input within flow-based models as shown in Fig.~\ref{fig:model_overview}(b), we adopt the open-source inpainting version of FLUX, called FLUX.Fill, into three application scenarios using task-specific LoRA: conditional generation, perception, and image editing.

\noindent(1) \textbf{Conditional generation tasks} involve creating new content from coarse input conditions, such as synthesizing photorealistic images from depth maps, canny edges, or segmentation maps, as well as performing colorization to generate plausible colors beyond the reference. AnyRefill utilizes these perceptual or grayscale references to produce coherent images aligned with the specified text prompt on the right canvas.

\noindent(2) \textbf{Perception tasks} focus on extracting perceptual information for image and scene understanding. In contrast to conditional generation, AnyRefill employs a reversed stitching direction to generate corresponding edge maps, depth maps, and segmentation results on the right canvas.

\noindent(3) \textbf{Image editing tasks} modifies existing content to improve quality or adjust specific attributes, such as deblurring, super-resolution, Ref-inpainting, and portrait modifications (age, gender, relighting). Age and gender editing require subtle adjustments to facial features while preserving background and clothing details. Relighting modifies foreground lighting effects based on background light direction and textual descriptions. Deblurring and super-resolution enhance image quality while maintaining scene consistency. Following LeftRefill, AnyRefill performs Ref-inpainting by using a left reference to fill missing regions with coherent structures.

\noindent\textbf{Discussion about More Tasks.} 
Despite their diverse objectives, all these tasks are seamlessly unified within the AnyRefill framework using the LPG formulation, demonstrating that it is a versatile approach for generative modeling. 
We present a representative and diverse set of vision tasks in this work, which are effectively addressed by AnyRefill. Moreover, we posit that our flexible LPG framework can also efficiently tackle numerous additional vision tasks, sharing a similar model design.
AnyRefill holds great potential for broader generalization in future research and applications.

\noindent\textbf{Discussion about More Prompt Images.}
As noted in~\cite{cao2024leftrefill}, LeftRefill supports multiple reference images, especially for novel view synthesis (NVS) and multi-view Ref-inpainting task. Given FLUX.Fill's attention-based architecture~\cite{peebles2023scalable}, we believe it has the potential to handle multi-view tasks. 
However, due to the model size and computational cost of FLUX series, these tasks are left for future exploration. 
In this work, we focus on broadening the scope of vision tasks that AnyRefill can address.
We present the superior performance of AnyRefill in various tasks in Sec.~\ref{sec:experiments} to prove its versatility and practical utility.

\subsection{Details in Learning AnyRefill \label{sec:detail_anyrefill} } 

\noindent\textbf{Curated Training Pairs for AnyRefill.}
The amount of training data is summarized in Tab.~\ref{tab:teaser_table}.
To empower AnyRefill with impressive generative capabilities, we fully leverage current state-of-the-art models to curate high-quality training data pairs.
For perceptual data involved in conditional generation and perception tasks, we construct tailored datasets using tools such as DepthAnything~\cite{yang2024depth} and GSAM~\cite{ren2024grounded,kirillov2023segment} for depth and segmentation maps respectively. We directly extract the Canny edge by employing OpenCV. Besides, grayscale-converted results from RGB images are regarded as natural training pairs for colorization.
In the image editing aspect, we utilize the open-source SD-based IC-Light~\cite{iclight2024} for relighting, while SeedEdit~\cite{shi2024seededit} is used for gender and age editing. 
To create degraded image pairs for deblurring, we add Gaussian noise to the images. For super-resolution, we achieve this by first downsampling the images and then upsampling them with the nearest interpolation strategy.
Additionally, all images are captioned by CogVLM2~\cite{hong2024cogvlm2} to obtain rich semantics with a thorough understanding of the image scene, enhancing the model's generative capabilities across diverse tasks.

\noindent\textbf{Task-specific LoRAs.}
We inject LoRA into all attention blocks of AnyRefill, covering linear layers for visual projection, text projection, and feed-forward layers. The formulation can be written as:
\begin{equation}\small
h = W_0x+B_{\phi}A_{\phi}x, \quad B_{\phi}\in\mathbb{R}^{d\times r},A_{\phi}\in\mathbb{R}^{r\times d},
\label{eq:lora}
\end{equation}
where $x$ and $h$ indicate input and output features with channels $d$ for the linear layers in attention blocks; $W_0$ denotes the frozen DiT's weights, while $B_{\phi},A_{\phi}$ are trainable low-rank matrices with much fewer parameters compared to $W_0$, \ie, $r \ll d$.
This enables AnyRefill to stably generate image content on the right canvas with the left reference, while preserving the ability to follow the instructions from text. The rectified flow loss function can be formulated as:
\begin{equation}\small
\mathcal{L}_{RF} = \mathbb{E}\left[ \| (\varepsilon - z_0) - v_\theta(\left[z_t;\hat{z}_0;\mathbf{M} \right], \phi, c_{txt}, t) \|^2 \right],
\label{eq:pt_objective}
\end{equation}
where $v_\theta$ is parameterized by a DiT model, $\phi$ denotes the trainable low-rank ‌matrices‌ of LoRA, and $c_{\textit{txt}}$ represents the textual semantics extracted by CLIP-L~\cite{radford2021learning} and T5~\cite{raffel2020exploring}. 

Benefiting from the extensibility of AnyRefill and LoRA, we can combine multiple task-specific LoRAs to handle more complex tasks. For example, integrating LoRA modules trained for age and gender editing allows the model to modify both attributes simultaneously in a cohesive and consistent manner, as illustrated in Fig.~\ref{fig:multi_lora_editing}. This modular design not only enhances the model’s flexibility but also demonstrates its potential to handle intricate tasks without additional training.

\noindent\textbf{Data Efficiency of AnyRefill.}
Reproducing high-quality data pairs from state-of-the-art models remains a notable challenge. To address this, we refer to the detailed ablation studies on canny-to-image generation in Sec.~\ref{Sec:Data_Efficiency_of_AnyRefill}, which analyzes the data efficiency of AnyRefill. The results demonstrate that, once the dataset size surpasses a specific threshold, AnyRefill achieves qualitative performance in editing tasks aligned with expectations.
Building on these insights, we empirically select the minimum number of training pairs, as outlined in Tab.~\ref{tab:teaser_table}. For tasks involving open-source models, we generate a moderate amount of pseudo-image pairs across different datasets. For the closed-source editing model SeedEdit, we manually curate 50 image pairs using its paid application. AnyRefill showcases exceptional distillation capabilities, efficiently leveraging a limited number of image pairs from closed-source or commercial models with minimal degradation in generative performance.
Our findings suggest that task-specific LoRA fine-tuning, guided by the AnyRefill LPG paradigm, can effectively adapt the flow-based model to new tasks using as few as dozens of image pairs.

\section{Experiments}
\label{sec:experiments}
\noindent\textbf{Datasets.}
For Ref-inpainting, we use image pairs from MegaDepth~\cite{li2018megadepth}, which includes many multi-view famous scenes collected from the Internet. To trade-off between the image correlation and the inpainting difficulty, we empirically retain image pairs with 40\% to 70\% co-occurrence with about 80k images and 820k pairs. The validation of Ref-inpainting also includes some manual masks from ETH3D scenes~\cite{schops2017multi} to verify the generalization. For the image-to-segment task, we generate images by FLUX and segment them by GSAM~\cite{ren2024grounded}. For other tasks, we construct the training dataset using DIV2K~\cite{agustsson2017ntire} and Flicker2K~\cite{lim2017enhanced}, both of which contain high-resolution images of diverse scenes and objects. The DIV2K dataset consists of 900 images, with 800 allocated for training and 100 for testing. The entire Flicker2K dataset, containing 2,650 images, is used solely for training. For image editing tasks, we curate 50 portrait images, either generated by FLUX or sourced from the Internet, and construct the training and testing sets using the method described in Sec.~\ref{section:task-specific}. All tasks are executed at a resolution of 512, while the LPG formulation is conducted in 512$\times$1024.

\noindent\textbf{Implementation Details.}
By default, we inherit most of the configurations from LeftRefill, with the key exception of the tuning method, \ie, LeftRefill's prompt tuning vs our task-specific LoRAs in Sec.~\ref{sec:detail_anyrefill}.
For the Ref-inpainting, 75\% masks are randomly generated, and 25\% are matching-based masks. For all other tasks, we masked the entire 512$\times$512 region of the right canvas for image synthesis. To adapt FLUX.Fill to various vision tasks, We employ LoRA adapters with a rank of 128. The AdamW optimizer is chosen with a learning rate of 1e-4 and a batch size of 16. The sampling step is set to 50 for better performance across different tasks.


\begin{figure}[h!]
\begin{center}
\includegraphics[width=1.0\linewidth]{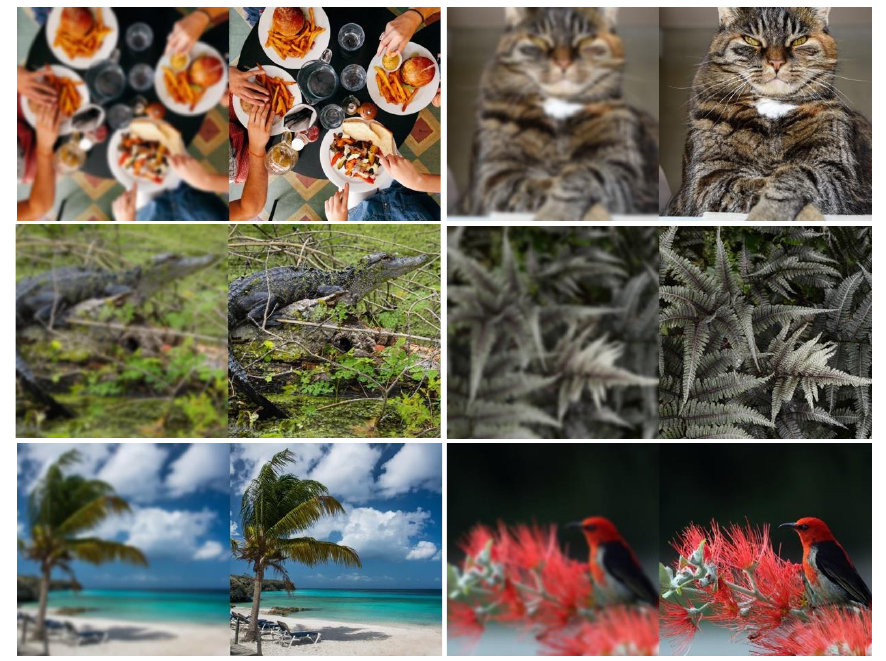}
\vspace{-0.2in}
\end{center}
   \caption{Qualitative results of the deblurring task. AnyRefill restores content and maintains consistency with the reference.}
   \label{fig:generation_deblurring}
\vspace{-0.1in}
\end{figure}

\begin{figure}[h!]
\begin{center}
\includegraphics[width=1.0\linewidth]{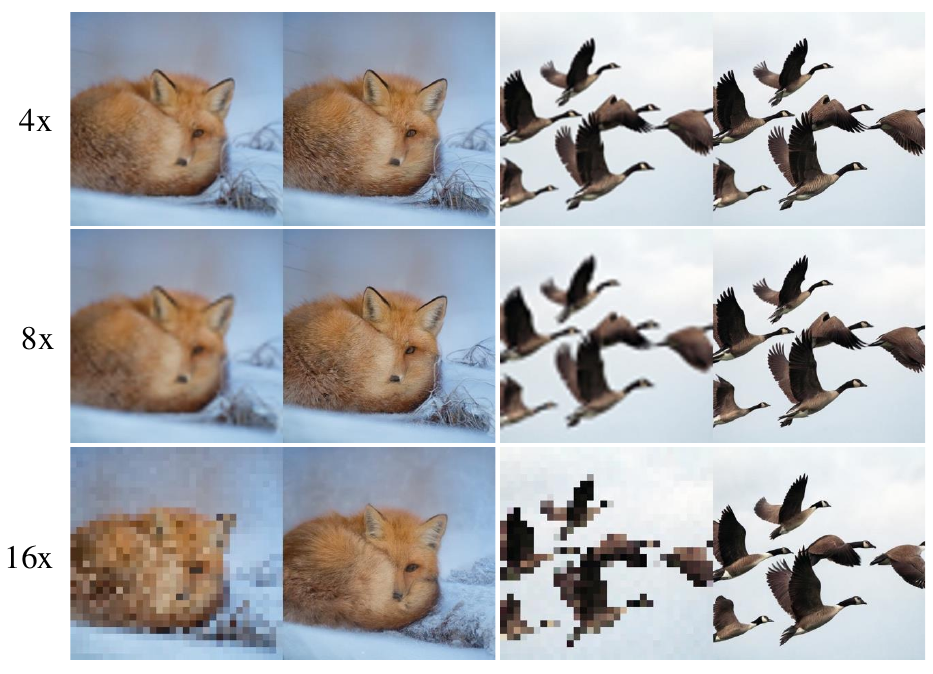}
\vspace{-0.2in}
\end{center}
   \caption{Qualitative results of super-resolution task across different upsampling ratios. AnyRefill generates high-fidelity images and maintains consistency with the reference.}
   \label{fig:generation_super_resolution}
\vspace{-0.2in}
\end{figure}

\subsection{Results of Image Editing Tasks.}
\begin{figure*}
\begin{center}
\includegraphics[width=0.95\linewidth, height=19.3cm]{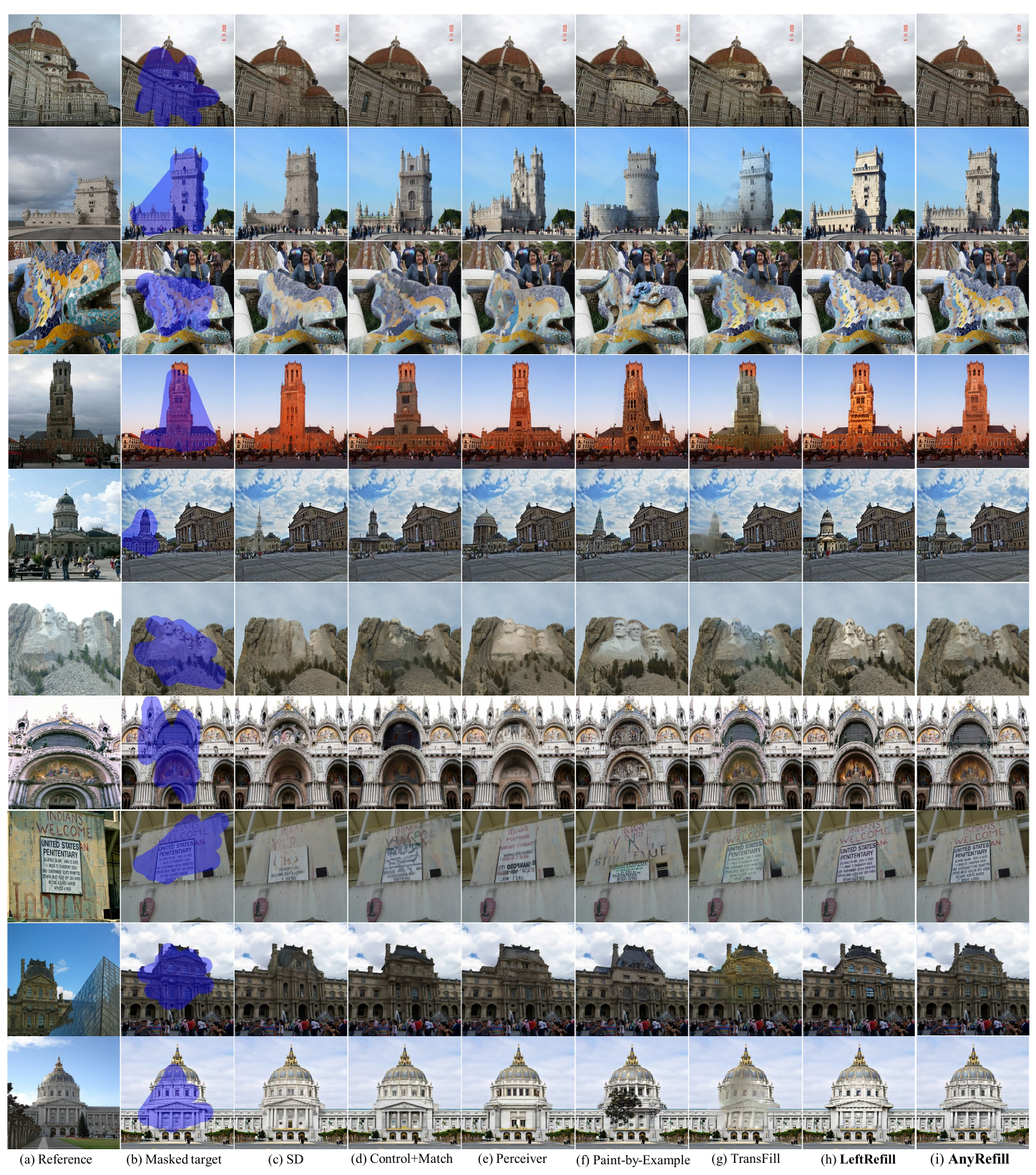}
\vspace{-0.15in}
\end{center}
   \caption{ Qualitative Ref-inpainting results on MegaDepth~\cite{li2018megadepth}. LeftRefill and AnyRefill exhibit superior performance compared to pioneer methods. 
   \label{fig:reference_guided_qualitative}}
\vspace{-0.15in}
\end{figure*}

\begin{table}[h]
\small 
\caption{Quantitative results for Ref-inpainting on MegaDepth~\cite{li2018megadepth} test set based on matching and manual masks. 
`ExParams’: the percentage of parameters required to be fine-tuned compared to base model. * means that the uncorrupted ground truth is visible for the matching. `No stitching': reference and target views are separate without spatial stitching, and only self-attentions are learned across them. AnyRefill achieves superior results compared with other state-of-the-art methods.
\label{tab:reference_inpainting_quantitative}}
\vspace{-0.125in}
\centering
\footnotesize
\renewcommand\tabcolsep{2pt}
\begin{tabular}{lccccl}
\hline
Methods & PSNR$\uparrow$ & SSIM$\uparrow$ & FID$\downarrow$ & LPIPS$\downarrow$ & ExParams\tabularnewline
\hline
SD (inpainting)~\cite{rombach2022high} & 19.841 & 0.819 & 30.260 & 0.1349 & +0\%\tabularnewline
FLUX.Fill~\cite{fluxfill2024} & 21.196 & 0.841 & 21.763 & 0.1204 & +0\%\tabularnewline
ControlNet~\cite{zhang2023adding} & 19.072 & 0.744 & 33.664 & 0.1816 & +42.3\%\tabularnewline
ControlNet+NewCrossAttn & 19.027 & 0.743 & 34.170 & 0.1805 & +53.9\%\tabularnewline
ControlNet+Matching{*}~\cite{tang2022quadtree} & 20.592 & 0.763 & 29.556 & 0.1565 & +42.3\%\tabularnewline
Perceiver+ImageCLIP~\cite{jaegle2021perceiver} & 19.338 & 0.745 & 32.911 & 0.1751 & +6.0\%\tabularnewline
Paint-by-Example~\cite{yang2023paint} & 18.351 & 0.797 & 34.711 & 0.1604 & +100.7\%\tabularnewline
TransFill~\cite{zhou2021transfill}(closed-source) & \textbf{22.744} & \textbf{0.875} & 26.291 & 0.1102 & --\tabularnewline
\hline
LeftRefill (no stitching) & 20.489 & 0.827 & 20.125 & 0.1085 & +$<$0.1\%\tabularnewline
LeftRefill \cite{cao2024leftrefill} & 20.926 & 0.836 & 18.680 & 0.0961 & +$<$0.1\%\tabularnewline
\rowcolor{gray!20}
\textbf{AnyRefill}  & \textbf{21.993} & \textbf{0.862} & \textbf{16.788} & \textbf{0.0945} & +3.0\%\tabularnewline
\hline 
\end{tabular}
\vspace{-0.05in}
\end{table}

\begin{figure*}
\begin{center}
\includegraphics[width=0.9\linewidth]{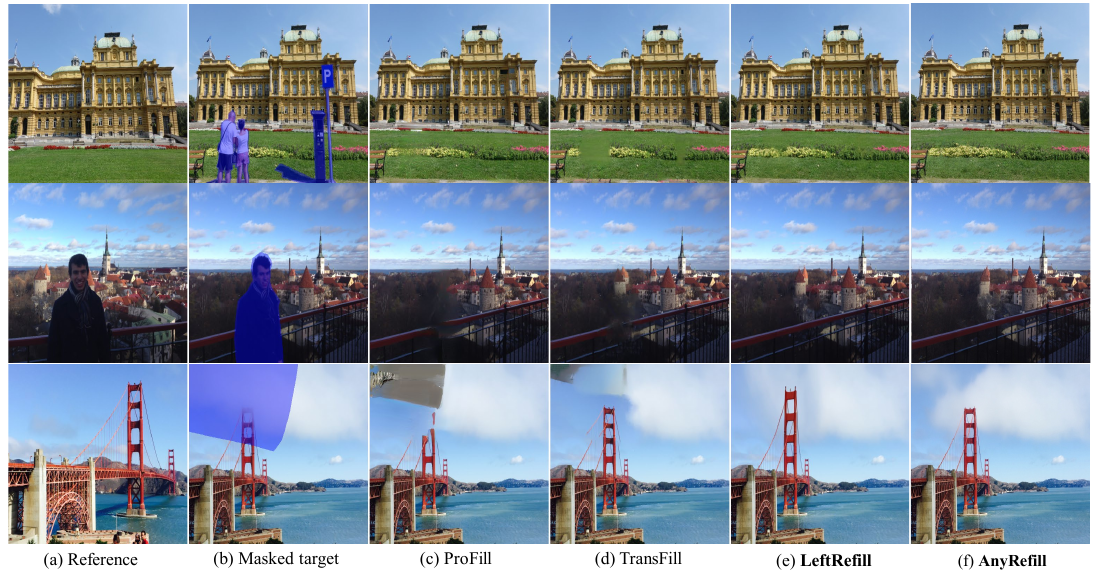}
\vspace{-0.15in}
\end{center}
   \caption{Qualitative Ref-inpainting results compared with ProFill~\cite{zeng2020high}, TransFill~\cite{zhou2021transfill}, LeftRefill~\cite{cao2024leftrefill}, and AnyRefill on the challenging real set provided by TransFill.
   \label{fig:real_set_qualitative}}
\vspace{-0.15in}
\end{figure*}

\label{sec:exp_image_editing}
\noindent\textbf{Ref-inpainting.}
\xieming{
We first thoroughly compared the specific Ref-inpainting method~\cite{zhou2021transfill} and existing image reference-based variants of SD in Tab.~\ref{tab:reference_inpainting_quantitative} and Fig.~\ref{fig:reference_guided_qualitative}. 
Note that ControlNet~\cite{zhang2023adding} fails to learn the correct spatial correlation between reference images and masked targets, even enhanced with trainable cross-attention learned between reference and target features. 
Furthermore, we try to warp ground-truth latent features with image matching~\cite{tang2022quadtree} as the reference guidance for ControlNet, but the improvement is not prominent.
Perceiver~\cite{jaegle2021perceiver} and Paint-by-Example~\cite{yang2023paint} align and learn image features from Image CLIP. Since image features from CLIP contain high-level semantics, they fail to deal with the fine-grained Ref-inpainting as shown in Fig.~\ref{fig:reference_guided_qualitative}(e)(f).
Though TransFill~\cite{zhou2021transfill} achieves proper results in PSNR and SSIM, it suffers from blur and color difference as in Fig.~\ref{fig:reference_guided_qualitative}(g) with challenging viewpoints.
AnyRefill enjoys substantial advantages in both qualitative and quantitative comparisons with moderate trainable weights, exhibiting superior capability compared with LeftRefill and other state-of-the-art methods. 
Particularly, spatially stitching reference and target views together achieves consistent improvements.
We further compare AnyRefill with TransFill on the officially provided real-world dataset in Fig.~\ref{fig:real_set_qualitative}. 
AnyRefill enjoys good generalization in unseen or occluded real-world scenes, because it gets rid of the constrained geometric warping from wrong 3D results.}

\begin{table}
    \centering
    \small
    \caption{Results of 4x super-resolution. AnyRefill easily switches between 4x, 8x, and 16x, while ESRGAN~\cite{wang2018esrgan} only provides an open-source 4x model, with AnyRefill showing competitive quantitative performance.}
    \begin{tabular}{c|c|ccc}
         \hline
         Upscaling & Methods & PSNR$\uparrow$ & SSIM$\uparrow$ & LPIPS$\downarrow$ \\
         \hline
         \multirow{3}{*}{4x}
         & ESRGAN & \underline{23.225} & 0.712 &  \textbf{0.138} \\
         & ESRGAN (PSNR) & \textbf{26.650} & \underline{0.817} &  0.243 \\
         & AnyRefill & 22.856 & \textbf{0.842} &  \underline{0.144} \\
         \hline
    \end{tabular}
    \label{tab:quantitative_super_resolution}
\vspace{-0.05in}
\end{table}

\noindent\textbf{Restoration.}
\xieming{Similar to right-canvas-based generation tasks, AnyRefill can also handle image restoration tasks, such as super-resolution and deblurring within the LPG formulation. For the deblur task applying Gaussian noise to the reference and the super-resolution task with nearest upsampling, AnyRefill demonstrates impressive performance in reconstructing fine details, effectively restoring high-quality content while maintaining consistency with the reference. We present the fine-tuning results in Fig.~\ref{fig:generation_deblurring}, Fig.~\ref{fig:generation_super_resolution} and Tab.~\ref{tab:quantitative_super_resolution}}

\begin{figure*}
\begin{center}
\includegraphics[width=1.0\linewidth]{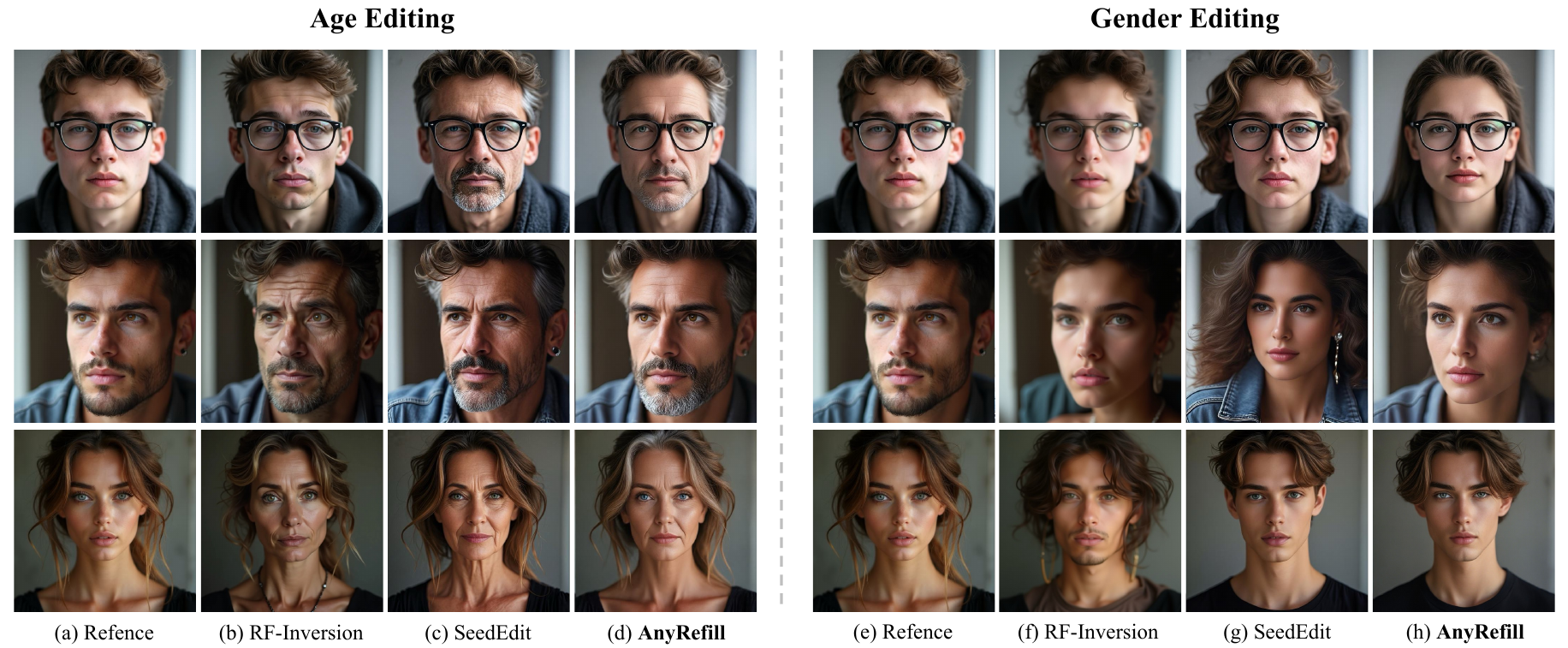}
\vspace{-0.2in}
\end{center}
   \caption{Qualitative results of age editing (left) and gender editing (right), which are compared among (b) RF-Inversion~\cite{rout2024semantic}, (c) SeedEdit~\cite{shi2024seededit}, (d) our AnyRefill with same prompts.}
   \label{fig:gender_editing_compare}
\vspace{-0.2in}
\end{figure*}

\begin{figure}
\begin{center}
\includegraphics[width=1.0\linewidth]{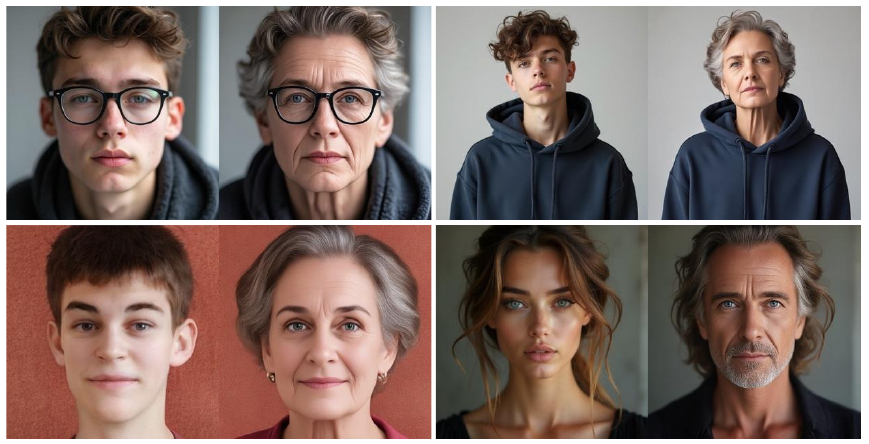}
\vspace{-0.2in}
\end{center}
   \caption{Qualitative results of AnyRefill with combined LoRAs. The left one denotes the original reference while the right one shows the edited result based on AnyRefill with both age and gender LoRAs.}
   \label{fig:multi_lora_editing}
\vspace{-0.2in}
\end{figure}

\begin{figure}[h!]
\begin{center}
\includegraphics[width=1.0\linewidth]{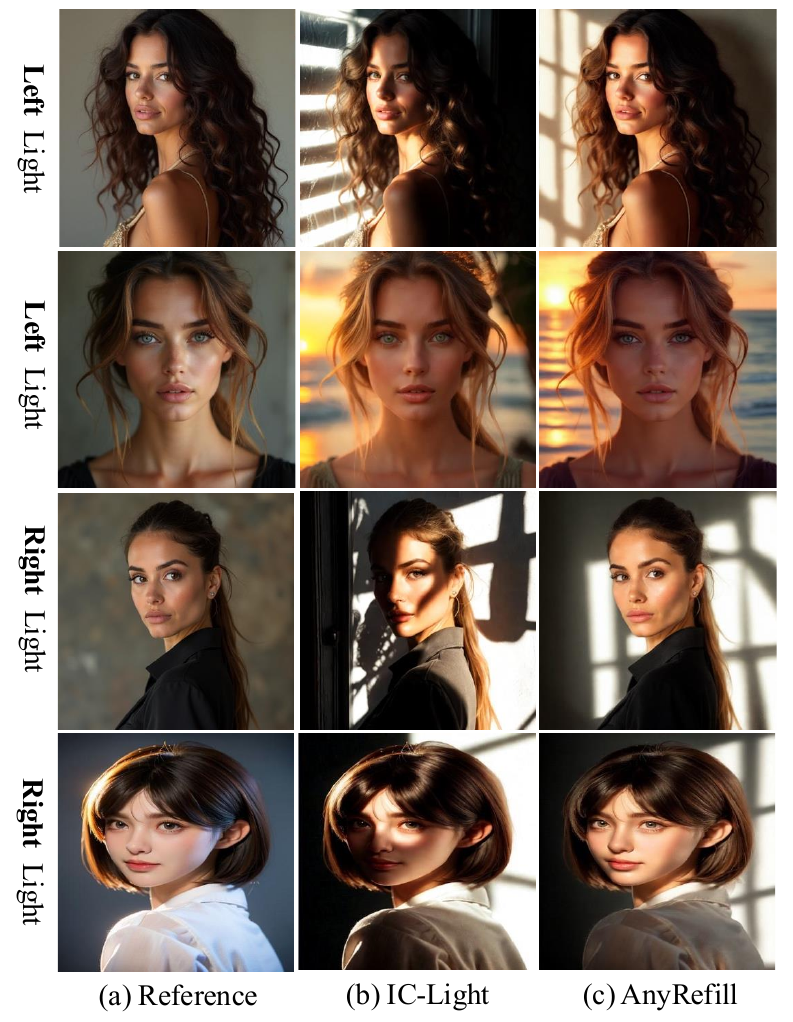}
\vspace{-0.2in}
\end{center}
   \caption{Qualitative results of relighting task. AnyRefill demonstrates strong facial preservation (zoom in for details) and shows competitive performance in lighting effects.}
   \label{fig:editing_relight}
\vspace{-0.1in}
\end{figure}

\begin{figure}
\begin{center}
\includegraphics[width=1.0\linewidth]{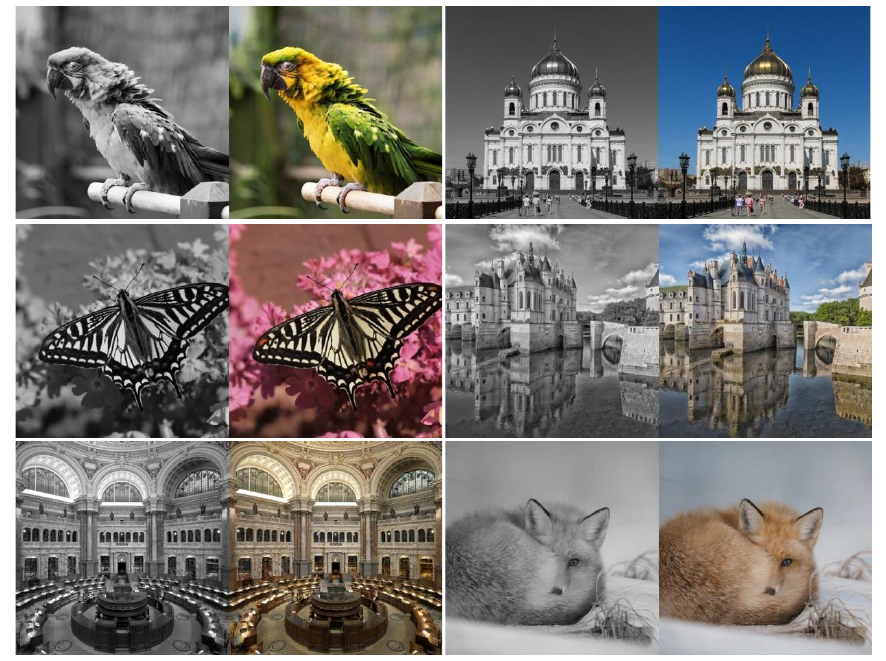}
\vspace{-0.2in}
\end{center}
   \caption{Qualitative results of colorization which require T2I generate harmonious scene.}
   \label{fig:colorize_editing}
\vspace{-0.1in}
\end{figure}

\noindent\textbf{Editing for Age, Gender, and Relighting.}
\xieming{For age and gender editing, we selected RF-Inversion~\cite{rout2024semantic}\footnote{\scriptsize RF-Inversion (commercial): \url{https://huggingface.co/spaces/rf-inversion/RF-inversion}.}, a tuning-free method, and SeedEdit~\cite{shi2024seededit}\footnote{\scriptsize SeedEdit (commercial): \url{https://jimeng.jianying.com/}.}, which involves a complex data pipeline and heavy data requirements, as comparative baselines — both of which are commercial state-of-the-art models. As shown in Fig.~\ref{fig:gender_editing_compare}, RF-Inversion generates text-aligned results but introduces noticeable stylistic changes. Additionally, RF-Inversion requires extensive manual parameter adjustments, leading to inconsistent results. It should be noticed that SeedEdit often produces slightly undesired modifications in clothing, which negatively impacts the overall editing quality. Thanks to AnyRefill’s strong alignment to the reference, it can generate high-quality, text-aligned editing results even with fine-tuning on a small number of image pairs. As shown in Fig.~\ref{fig:gender_editing_compare}, AnyRefill consistently outperforms RF-Inversion and achieves results comparable to SeedEdit in gender editing tasks. Furthermore, Fig.~\ref{fig:multi_lora_editing} highlights AnyRefill’s flexibility in handling complex scenarios. By combining different LoRA parameters, it can simultaneously edit multiple attributes.
For the relighting task, as shown in Fig.~\ref{fig:editing_relight}, AnyRefill achieves results on par with IC-Light~\cite{iclight2024} while avoiding its complicated data construction pipeline by training LoRAs from different light directions. Notably, in the editing tasks, AnyRefill requires only 50 image pairs for age and gender editing, while 35 image pairs are used for relighting.}

\begin{figure*}
\begin{center}
\includegraphics[width=1.0\linewidth]{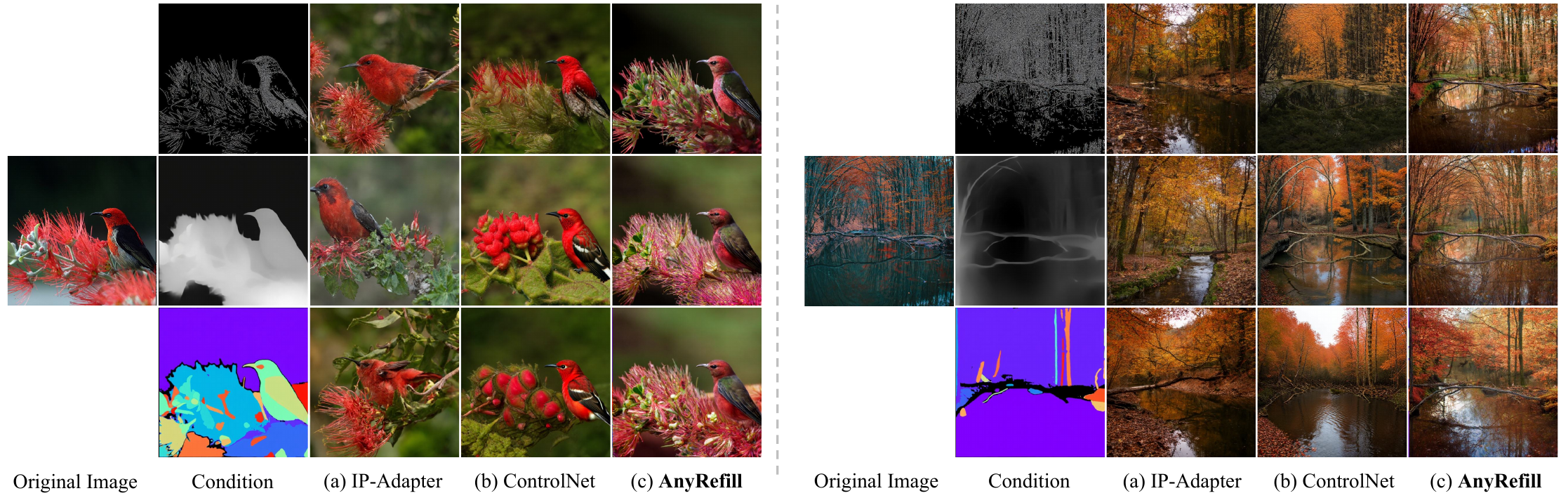}
\vspace{-0.15in}
\end{center}
   \caption{Qualitative results of generation tasks based on canny edge, depth map, and segmentation. AnyRefill generates results that align closely with the reference content in generation tasks.
   \label{fig:generation_task_results}}
\vspace{-0.2in}
\end{figure*}

\subsection{Results of Other Tasks}

We also provide detailed qualitative and quantitative evaluations for various conditional generation and perception tasks mentioned in this paper. 
Note that all these tasks are classical vision tasks, and we include the results to show that our AnyRefill can address them in one framework with few training data.

\begin{table}
    \centering
    \small
    \caption{Quantitative comparison of different image condition injection methods under limited training data scenarios (100 training pairs) in conditional generation tasks.}
    \begin{tabular}{c|c|cc}
         \hline
         Condition & Methods & CLIP$\uparrow$ & LPIPS$\downarrow$ \\
         \hline
         \multirow{3}{*}{Depth}
         & FLUX+IP-Adapter & 0.8505 &  0.713 \\
         & FLUX+ControlNet & 0.8650 &  0.617 \\
         & AnyRefill & \textbf{0.8828} &  \textbf{0.593} \\
         \hline
         \multirow{3}{*}{Canny}
         & FLUX+IP-Adapter & 0.8572 &  0.718 \\
         & FLUX+ControlNet & 0.8607 &  0.579 \\
         & AnyRefill & \textbf{0.8878} &  \textbf{0.547} \\
         \hline
         \multirow{3}{*}{Segment}
         & FLUX+IP-Adapter & 0.8555 & 0.736  \\
         & FLUX+ControlNet & 0.8538 & 0.638  \\
         & AnyRefill & \textbf{0.8696} &  \textbf{0.601} \\
         \hline
    \end{tabular}
    \label{tab:quantitative_conditional_generation}
\vspace{-0.05in}
\end{table}

\noindent \textbf{Results of Conditional Generation Tasks}.
We use ControlNet and IP-Adapter as comparative baselines for our experiments. All models are trained with 100 data pairs to ensure fairness, although AnyRefill can be effectively trained with just 10 pairs, as shown in Tab.~\ref{tab:teaser_table}. These widely adopted image condition injection methods demonstrate strong reference alignment capabilities and can handle various conditional generation tasks, making them reasonable competitors to AnyRefill.
As shown in Fig.~\ref{fig:generation_task_results}, AnyRefill exhibits impressive capabilities in reference-guided synthesis, seamlessly generating the full right canvas based on the left-side condition. Notably, AnyRefill outperforms both ControlNet and IP-Adapter in terms of reference alignment and image quality.
Given any image condition, AnyRefill can effectively leverage both the reference and the prompt to generate contextually appropriate content with limited training data. The colorization task, as depicted in Fig.~\ref{fig:colorize_editing}, further highlights AnyRefill’s strong text alignment capabilities while preserving the overall harmony of the scene.
In addition, we provide quantitative comparisons using CLIP-Score and LPIPS to evaluate the quality of the generated images across different criteria, including depth, canny edge, and segmentation, in a few-shot setting. Leveraging the LPG formulation to inject visual context, AnyRefill consistently outperforms ControlNet and IP-Adapter, as summarized in Tab.~\ref{tab:quantitative_conditional_generation}.
More results can be found in the supplementary.

\begin{figure}
\begin{center}
\includegraphics[width=1.0\linewidth]{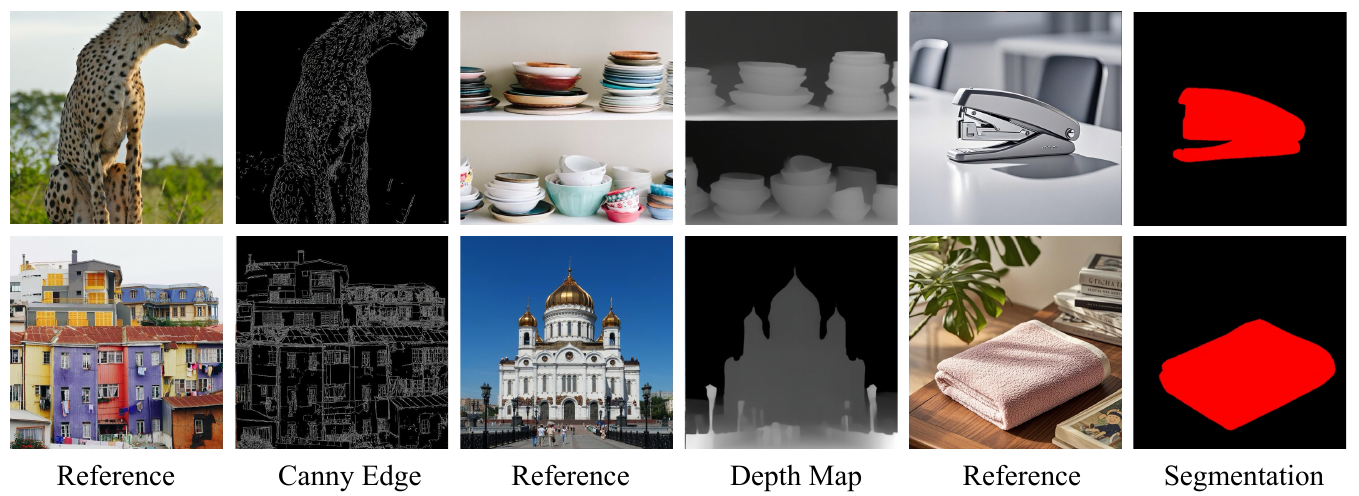}
\end{center}
   \caption{Qualitative results of perception tasks, including canny extraction, depth map generation, and foreground segmentation. AnyRefill demonstrates versatility across different tasks.
   \label{fig:generation_perception}}
\vspace{-0.2in}
\end{figure}

\noindent \textbf{Results of Perception Tasks}.
We extend AnyRefill to include perception tasks, such as depth map generation, edge map generation, and segmentation. While we acknowledge that each of these tasks has its own sub-research community within the vision field, the primary goal of incorporating these tasks is to demonstrate that AnyRefill is capable of addressing them in an ``all-in-one'' framework. We do not claim that AnyRefill achieves state-of-the-art performance on each of these tasks. In fact, many previous works~\cite{ke2023repurposing, zhao2023unleashing}, by leveraging larger training datasets and `bespoke' architectures, can achieve top-tier performance on these benchmarks. However, such models typically cannot solve these tasks in an integrated ``all-in-one'' fashion as AnyRefill does.
The visualized results for these tasks are presented in Fig.~\ref{fig:generation_perception}. The depth map illustrates the spatial relationships between objects in the scene, clearly depicting the relative distances of elements like the “lamp post” and “castle.” Edge extraction and segmentation highlight the model’s ability to perceive and isolate critical details. For example, in the foreground segmentation task, objects such as the “stapler” and “towel” are accurately segmented from a complex background, showcasing AnyRefill’s versatility and ability to handle a variety of tasks within a unified framework.

\subsection{Ablation Studies and Analysis}
\label{sec:ablation}

\begin{figure}
\begin{center}
\includegraphics[width=0.95\linewidth]{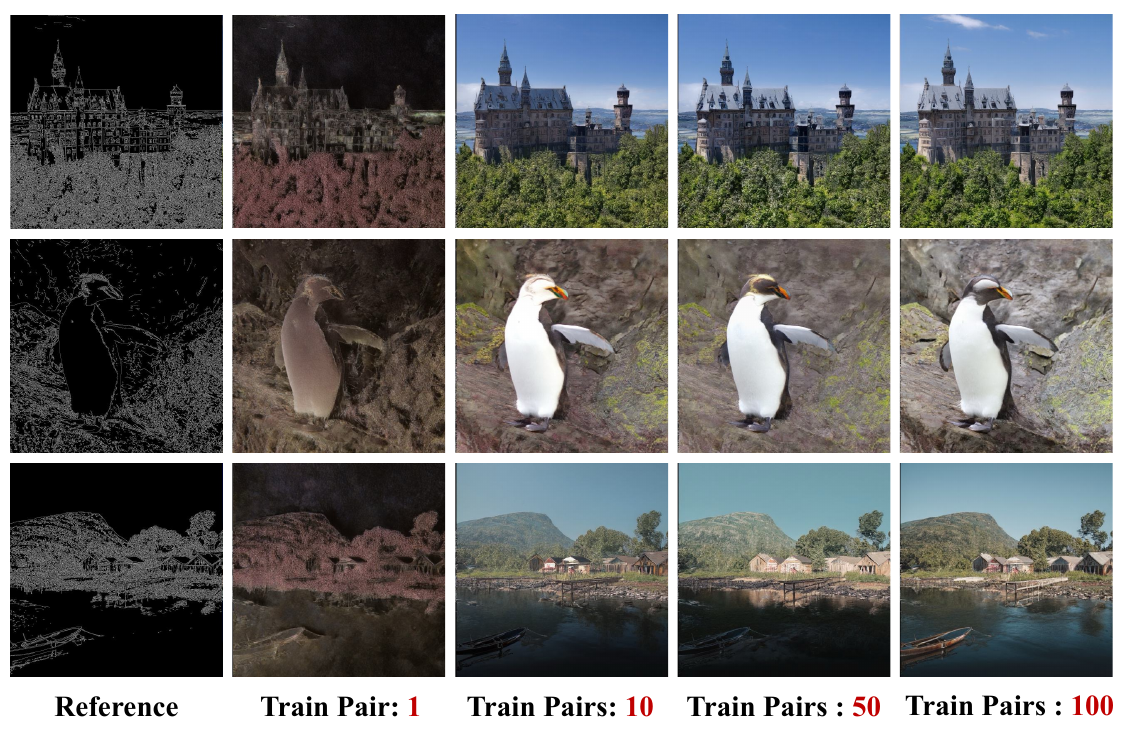}
\vspace{-0.2in}
\end{center}
   \caption{Ablation study of training pairs. AnyRefill can learn the basic pattern from limited data, with generation quality improving as data quantity increases. 
   \label{fig:ablation_train_pairs}}
\vspace{-0.2in}
\end{figure}

\noindent\textbf{Data Efficiency of AnyRefill.} \label{Sec:Data_Efficiency_of_AnyRefill}
\xieming{As previously mentioned, AnyRefill requires only a few dozen image pairs for effective training. To further investigate this, we explore the lower bound of training data required for the LPG formulation in rectified flow-based models. Using the generation task as an example, we discover that when the model is trained repeatedly on just a single stitched image, it surprisingly learns the basic LPG pattern of generating the right canvas from the left reference. However, the resulting image quality and color accuracy are significantly poor. When the training dataset is increased to 10 image pairs, we observe a noticeable improvement in color accuracy, though the overall image quality remains suboptimal. Further increasing the dataset to hundreds even thousands of images leads to gradual improvements in detail and overall quality, as shown in Fig.~\ref{fig:ablation_train_pairs}. We assume that the amount of training data affects different aspects of the model’s performance, such as pattern learning, image quality, and text alignment. For instance, aligning with complex textual prompts requires a substantial number of image pairs and multiple training iterations. We will further discuss this relationship in future work.}

\begin{figure}
\begin{center}
\includegraphics[width=0.95\linewidth]{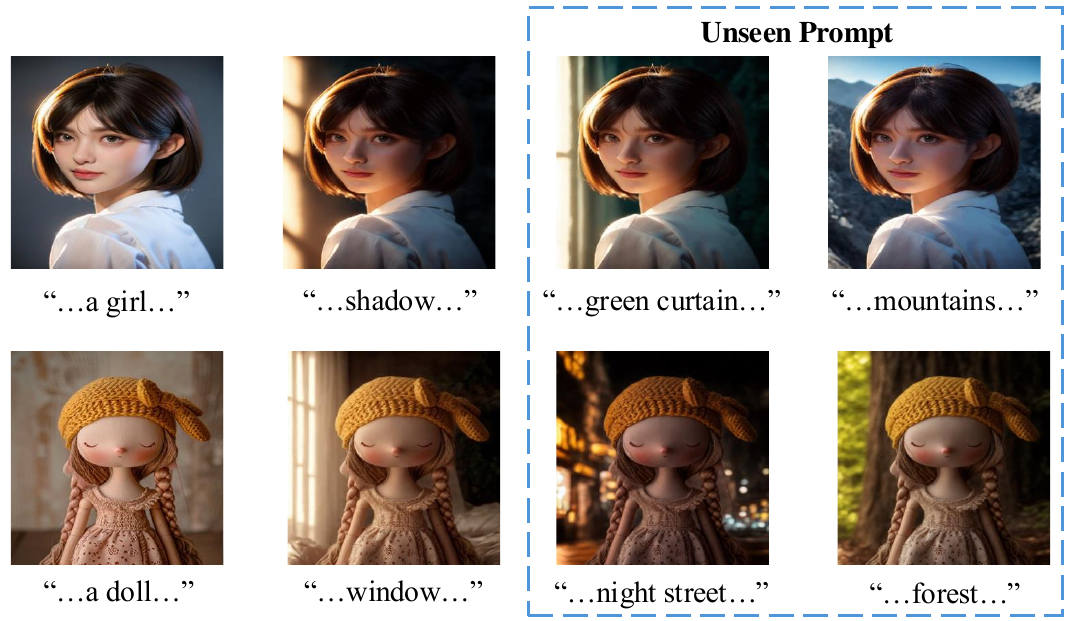}
\vspace{-0.1in}
\end{center}
   \caption{The textual alignment capability AnyRefill with unseen prompt, proving AnyRefill does not overfit to the training text.
   \label{fig:prompt_follow_ablation}}
\vspace{-0.15in}
\end{figure}

\noindent\textbf{Textual Alignment of AnyRefill. }\xieming{To evaluate the textual alignment capability of AnyRefill and ensure it does not overfit the small training dataset, we conduct experiments using unseen textual descriptions. Taking the relighting editing task as an example, we introduce new elements into the prompts, as illustrated in Fig.~\ref{fig:prompt_follow_ablation}. The results show that AnyRefill successfully maintains the consistency of the foreground while modifying the background according to the prompt. Furthermore, it accurately adjusts the lighting direction as instructed, demonstrating its ability to generalize to novel textual inputs beyond the training data with T2I model priors.}

\begin{figure}
\begin{center}
\includegraphics[width=0.95\linewidth]{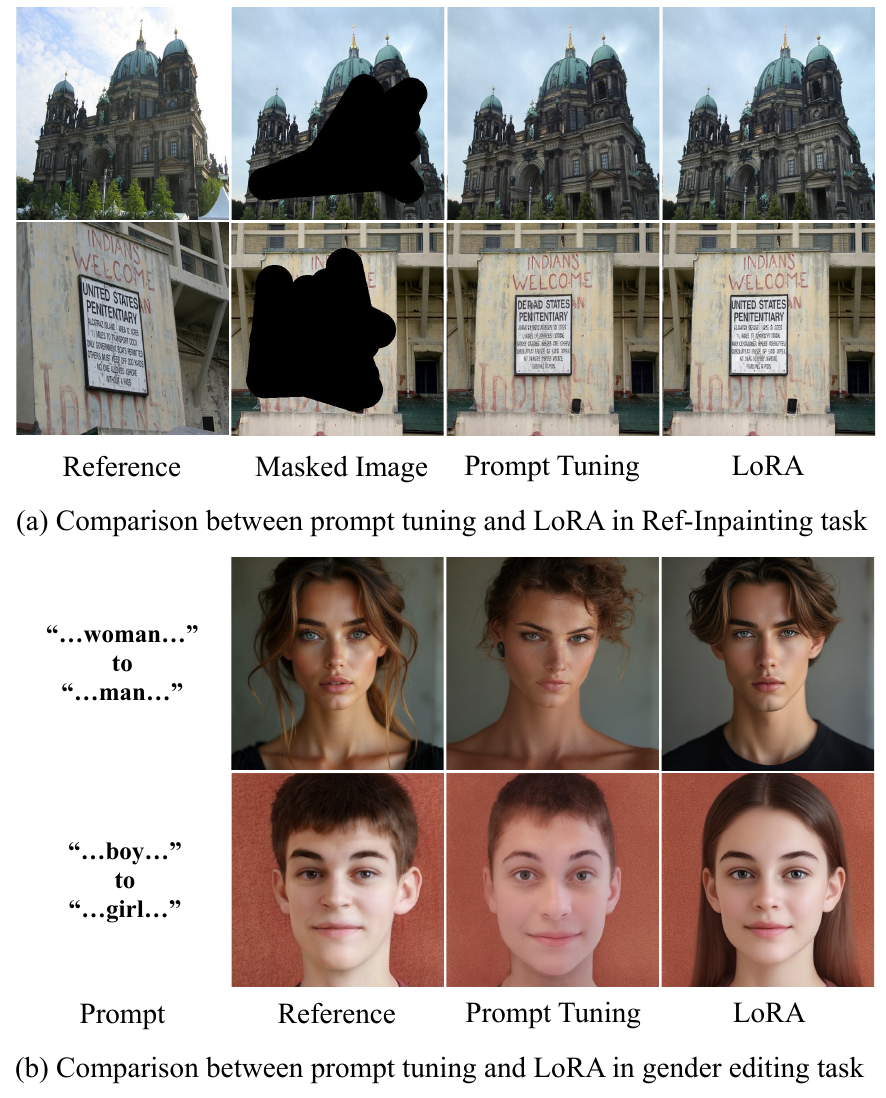}
\vspace{-0.2in}
\end{center}
   \caption{Visualization results of AnyRefill with Prompt Tuning and LoRA. AnyRefill with prompt tuning shows little degradation (textual recovery) in Ref-inpainting compared to LoRA (a), but performs poorly in editing tasks with bad identity preserving (b).
   \label{fig:anyRefill_with_prompt_tuning}}
\end{figure}

\begin{figure}[h!]
\begin{center}
\includegraphics[width=0.95\linewidth]{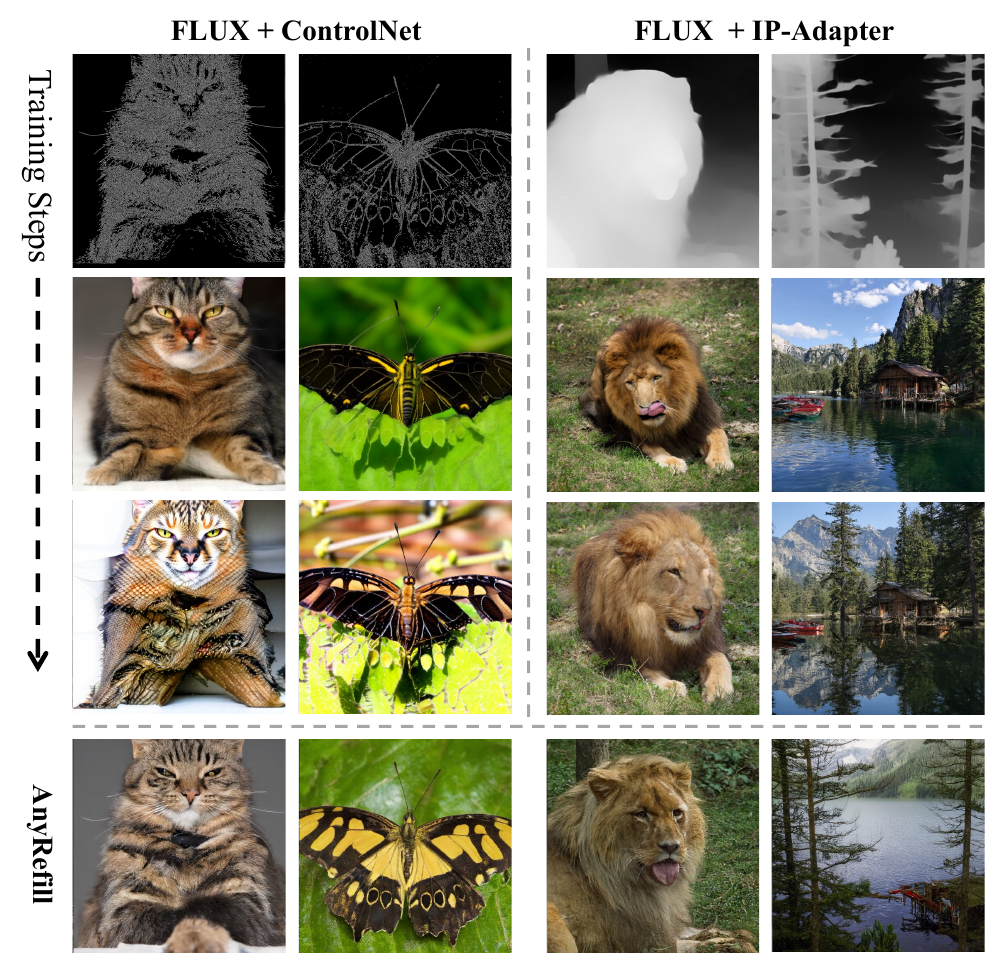}
\vspace{-0.2in}
\end{center}
   \caption{Visualization results of FLUX+ControlNet and FLUX+IP-Adapter under limited training data pairs. Compared to AnyRefill (LoRA+LPG-inpainting), FLUX+ControlNet suffers from model collapse while training with more steps. FLUX+IP-Adapter fails to precisely align with the reference.
   \label{fig:ablation_controlnet_ipadapater}}
\vspace{-0.25in}
\end{figure}

\noindent\textbf{AnyRefill with Prompt Tuning.} \xieming{Reflecting on recent state-of-the-art T2I models, such as Stable Diffusion 3~\cite{esser2024scaling}, which leverage large language models for precise textual semantics and image-text alignment, we evaluate AnyRefill, based on text-driven FLUX, under prompt tuning paradigm.  We inject semantic information into FLUX’s CLIP branch using 50 learnable token embeddings, following LeftRefill’s method, while the T5 branch keeps frozen. As shown in Fig.~\ref{fig:anyRefill_with_prompt_tuning}, we conduct experiments on Ref-inpainting and gender editing using AnyRefill without LoRA. In the Ref-inpainting task, the model effectively uses contextual inpainting capability to extract content from the reference image with minimal impact from fine-tuning. However, beyond extracting content from reference images, editing tasks require adjusting image attributes, such as facial appearance, based on editing details derived from text. Therefore, when CLIP fails to capture the diversity in text editing details—the sole source of control for editing tasks—the performance of AnyRefill with prompt tuning degrades significantly. This confirms the critical role of task-specific LoRA in enabling AnyRefill to adapt to a wide range of vision tasks.
}

\noindent\textbf{Comparing to ControlNet~\cite{zhang2023adding} and IP-Adapter~\cite{ye2023ip}.\label{sec:ablation_controlnet_ipadapter}} \xieming{We use Alimama’s FLUX-ControlNet~\cite{alimamacontrolnet2024} as the pre-trained model and fine-tune it to evaluate its performance on various tasks under limited data pairs. During the early stages of fine-tuning, we observe that ControlNet demonstrates relatively high image generation quality but struggles to follow the reference. As the model converges, the generated image quality significantly degrades. This indicates that ControlNet faces a trade-off between reference following and image quality under data limitation. Since there is currently no pre-trained IP-Adapter model available for the FLUX, we re-train the adapter module from scratch. IP-Adapter consistently generates misaligned images compared to the reference throughout the entire training process under limited data pairs. The experimental results are shown in Fig.~\ref{fig:ablation_controlnet_ipadapater}.}

\noindent\textbf{Self-Attention Analysis. }
\begin{figure}
\begin{center}
\includegraphics[width=0.98\linewidth]{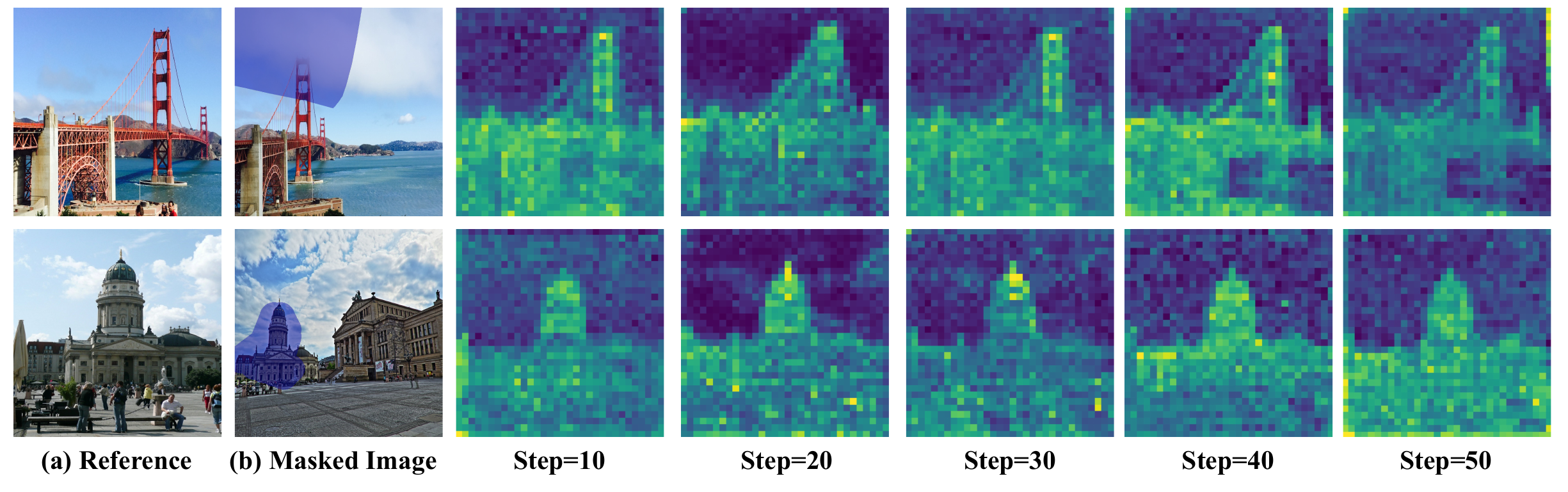}
\vspace{-0.1in}
\end{center}
   \caption{ Visualization of attention scores from reference views in AnyRefill for Ref-inpainting task across different Rectified Flow steps.
   \label{fig:attention_map_vis}}
\vspace{-0.2in}
\end{figure}
We visualize the attention scores of reference images in Ref-inpainting every 10 RF sampling steps, as shown in Fig.~\ref{fig:attention_map_vis}. By the 10th sampling step, the shape of the landmarks is already clearly visible. This demonstrates that the LPG input structure facilitates the self-attention modules in correctly attending to relevant regions of the left-side reference image during the initial stages of the sampling process, serving as evidence of AnyRefill achieving state-of-the-art performance. Furthermore, as the sampling steps progress, AnyRefill progressively refines the structure without introducing any drift.

\section{Conclusion}
In this paper, we introduce the Left-Prompt-Guided (LPG) formulation, inspired by the intuitive workflow of human painters. This approach spatially combines the reference and target images into a contextual inpainting task. Building on this foundation, we present AnyRefill, a framework leveraging a rectified flow-based DiT model to address diverse vision tasks—such as conditional generation, perception, and image editing—as LPG-inpainting tasks in an end-to-end manner, even with very limited training pairs.
By utilizing task-specific LoRAs and the robust attention mechanisms inherent in large text-to-image (T2I) models, AnyRefill achieves efficient and versatile performance across these tasks. Extensive experiments validate the effectiveness and efficiency of the proposed AnyRefill framework.

\bibliographystyle{IEEEtran}
\bibliography{main}

\newpage
\appendix
\input{supplementary}

\end{document}

%% file: supplementary.tex
\subsection{Broader Impacts}
This paper exploited image synthesis with text-to-image models. Because of their impressive generative abilities, these models may produce misinformation or fake images. 
So we sincerely remind users to pay attention to it.
Besides, privacy and consent also become important considerations, as generative models are often trained on large-scale data.
Furthermore, generative models may perpetuate biases present in the training data, leading to unfair outcomes. 
Therefore, we recommend users be responsible and inclusive while using these text-to-image generative models.
Note that our method only focuses on technical aspects. All pre-trained models used in this paper are all open-released.

\subsection{Supplementary Preliminaries}
\label{sec:data_processing}

\begin{figure}[h!]
\begin{center}
\includegraphics[width=1.0\linewidth]{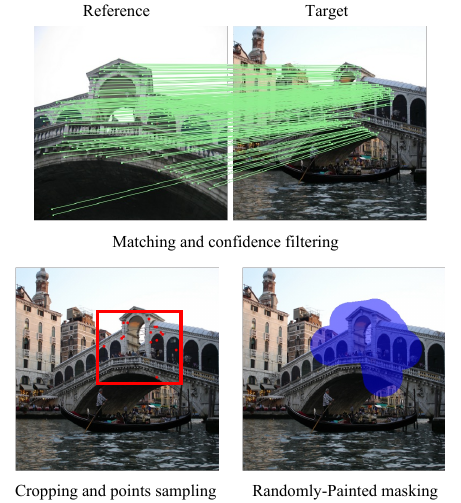}
\vspace{-0.15in}
\end{center}
   \caption{ The illustration of matching-based masking for Ref-inpainting task.
   \label{fig:data_process}}
\vspace{-0.15in}
\end{figure}

\label{sec:preliminary_anyrefill}

\noindent\textbf{Preliminaries of FLUX: Rectified Flow (RF)~\cite{liu2022flow}.}
Generative models aim to learn a mapping from a noise distribution $p_1$ to a data distribution $p_0$, where $p_0$ typically represents real-world data such as images or videos, and $p_1$ is often chosen as a standard Gaussian distribution. RF defines a straightforward approach to bridge these two distributions by constructing a straight trajectory in the latent space. This is achieved by modeling a time-dependent flow governed by  Ordinary Differential Equation (ODE). 

\begin{equation}
    {dZ_t} = v_\theta(Z_t, t){dt}, \quad t \in [0, 1]
\end{equation}
where $Z_t \in p_t$ represents the intermediate distribution at time $t$ and the velocity field $v(Z_t, t)$ is parameterized by a neural network $v_\theta$.
The forward process in rectified flow linearly interpolates between real data $X_0 \sim p_0$ and Gaussian noise $X_1 \sim p_1$. At each timestep $t$, the interpolated sample is defined as:
\begin{equation}
    X_t = (1 - t) X_0 + t X_1.
\end{equation}
This simple linear combination ensures that the data progressively transitions from  $X_0$ at $t=0$ to $X_1$ at $t=1$. The differential form of this interpolation is given by $dX_t = (X_1 - X_0) dt$.
To learn the velocity field, the network $v_\theta$ is trained to approximate the velocity between $X_0$ and $X_1$ along the interpolated path:
\begin{equation}
    \mathcal{L} = \left[ \| (X_1 - X_0) - v_\theta(X_t, t, c) \|^2 \right] dt,
\end{equation}
where $c$ is a text prompt condition in T2I flow-based models.
The sampling process in RF involves solving the ODE in reverse, starting from a Gaussian noise sample  $Z_N \sim \mathcal{N}(0, I)$. A sequence of timesteps $\{t_N, \ldots, t_0\}$ is defined to iteratively generate real data distribution samples:
\begin{equation} 
    Z_{t_{i-1}} = Z_{t_i} + (t_{i-1} - t_i) v_\theta(Z_{t_i}, t_i, c),
\end{equation}
where $i$ runs from $N$ to 0. 
\IEEEpubidadjcol

\label{sec:matching_mask}

\noindent\textbf{Data Processing for Ref-inpainting: Matching-based Masking.}
We follow LeftRefill~\cite{cao2024leftrefill} to conduct matching-based masking.
For the Ref-inpainting, we find that the widely used irregular mask~\cite{dong2022incremental,zhou2021transfill,zhao2022geofill} fails to reliably evaluate the capability of spatial transformation and structural preserving. Therefore, as shown in Fig.~\ref{fig:data_process}, we propose the matching-based masking method.
Specifically, we first utilize the scene info provided by MegaDepth~\cite{li2018megadepth} to select out the image pairs which have an overlap rate between 40\% and 70\% Second, for each image pair, we use a feature matching model~\cite{tang2022quadtree} to detect matching key-points between the images and assign each key-points pair a confidence score. 
Next, we filter out those pairs with low confidence scores with the threshold of 0.8. Then we randomly crop a 20\% to 50\% sub-space in the matched region and sample 15 to 30 key points as vertices to be painted across for the final masks. 
The matching-based mask not only improves the reliability during the evaluation but also facilitates the performance.

We split 505 pairs from MegaDepth~\cite{li2018megadepth} as the validation, including some manual masks from ETH3D scenes~\cite{schops2017multi}.

\subsection{Supplemental Experimental Results}
\begin{figure*}[htb!]
\begin{center}
\includegraphics[width=1.0\linewidth]{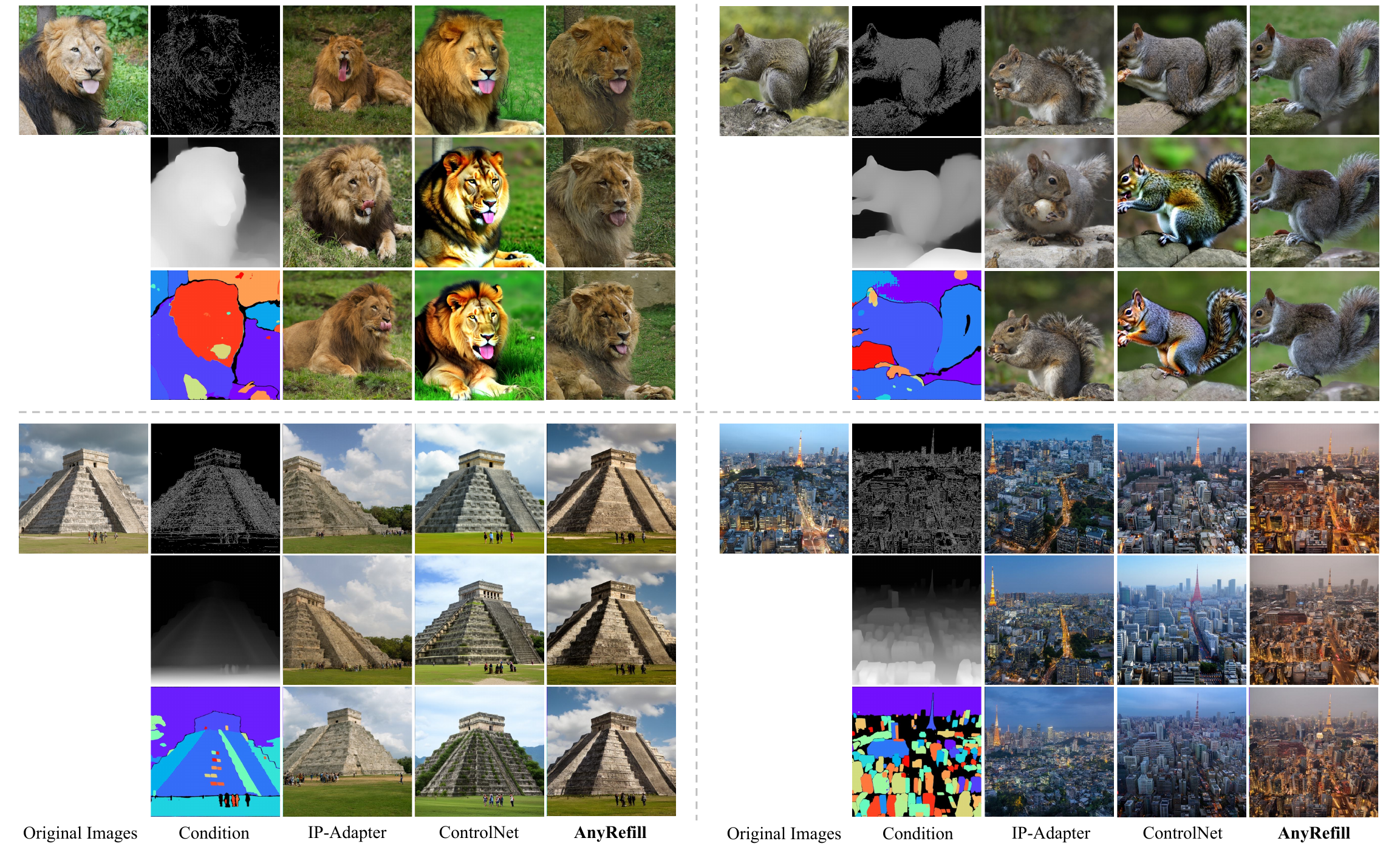}
\end{center}
   \caption{ The illustration of conditional generation tasks results, including canny-to-image, depth-to-image, and segment-to-image.
   \label{fig:supp_more_generation}}
\vspace{-0.2in}
\end{figure*}
\noindent\textbf{More Conditional Generation Results.} We show more impressive results of AnyRefill in Fig.~\ref{fig:supp_more_generation}. We selected the model output of ControlNet before its collapse as the comparison result. As shown in the figure, AnyRefill generates more realistic and reference-aligned images. While ControlNet also demonstrates decent alignment capabilities, IP-Adapter performs poorly due to data limitations, producing results that are entirely misaligned with the reference.

\noindent\textbf{More Editing Results.}
\begin{figure*}
\begin{center}
\includegraphics[width=1.0\linewidth]{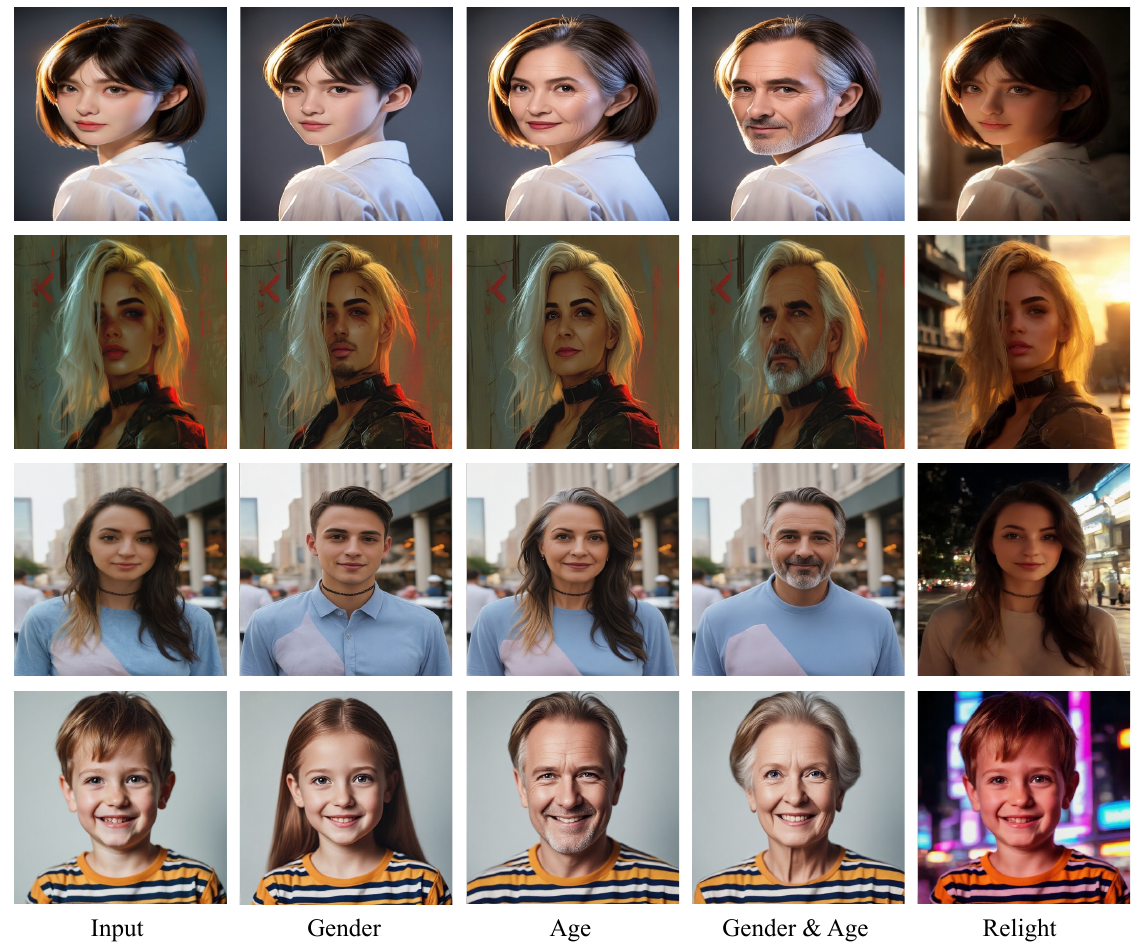}
\end{center}
   \caption{ The illustration of image editing tasks, including gender editing, age editing, and relighting.
   \label{fig:supp_more_editing}}
\vspace{-0.2in}
\end{figure*}
We show more image editing results in Fig.~\ref{fig:supp_more_editing} to verify that AnyRefill is a unified framework across various challenging tasks. 
AnyRefill seamlessly switches between different task-specific LoRAs, unifying highly challenging image editing tasks such as age editing, gender editing, and relighting under a single architecture, as shown in Fig.~\ref{fig:supp_more_editing}. In these qualitative results, AnyRefill preserves the foreground to the greatest extent while adjusting editing attributes based on textual input, achieving impressive outcomes.

\noindent\textbf{More Perception Results.}
\begin{figure*}[htb!]
\begin{center}
\includegraphics[width=1.0\linewidth]{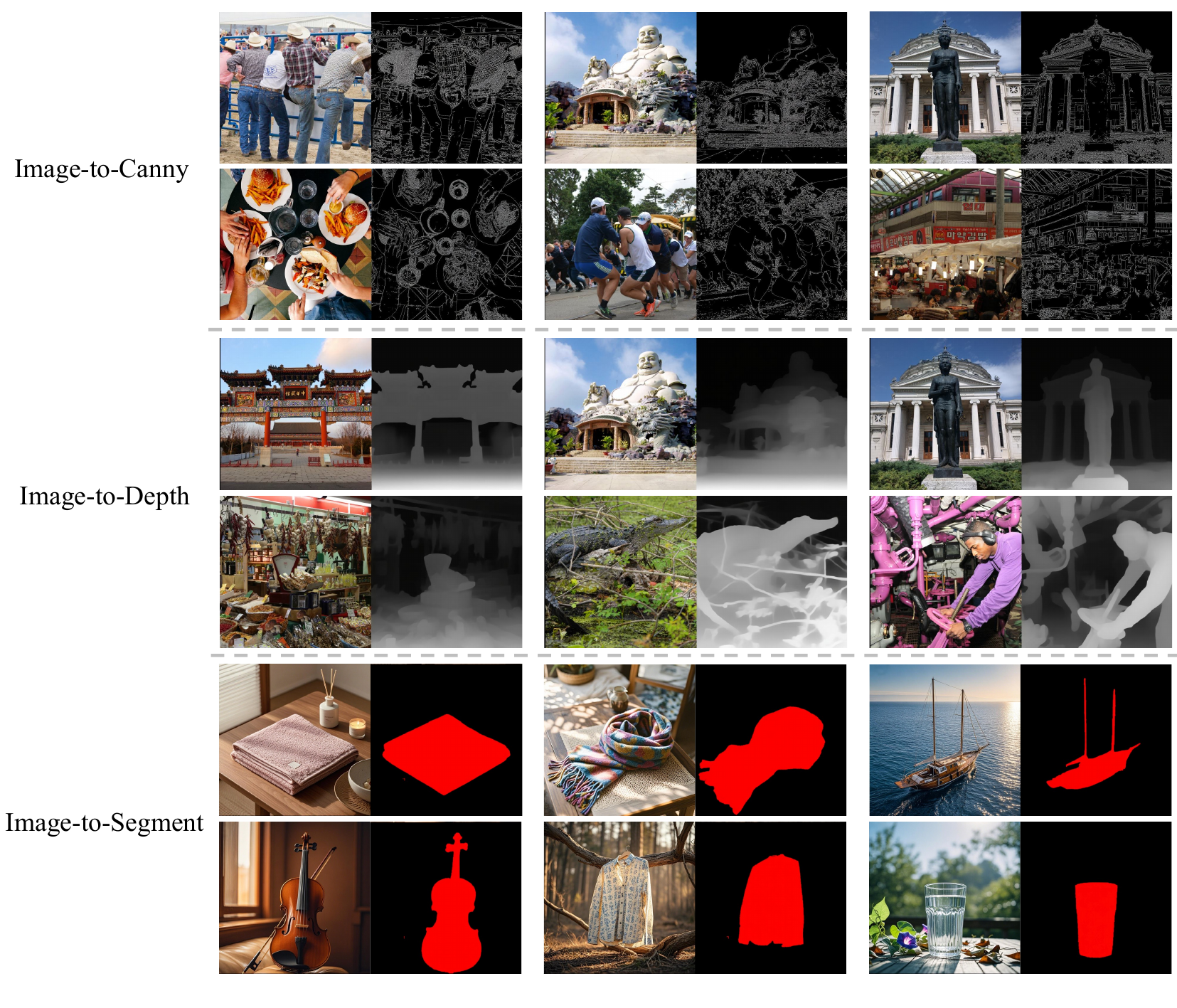}
\end{center}
   \caption{ The illustration of perception tasks, including image-to-canny, image-to depth, and image-to- segment.
   \label{fig:supp_more_perception}}
\vspace{-0.2in}
\end{figure*}
We show more visual perception results of AnyRefill in Fig.~\ref{fig:supp_more_perception}. AnyRefill can perform Image-to-Canny, Image-to-Depth, and Image-to-Segment tasks without requiring any modifications to the model architecture. As shown in the figure, AnyRefill demonstrates precise spatial information extraction, validating its potential to extend to a wider range of perception tasks.

\subsection{Inference Speed}
\begin{table}
\small 
\caption{Inference speed of AnyRefill under 50 RF sampling steps.
\label{tab:infer_speed}}
\vspace{-0.1in}
\centering
\begin{tabular}{cc}
\hline
Input size & Sec/image \tabularnewline
\hline
512$\times$512 & $\sim$ 5.26  \tabularnewline
\hline
512$\times$1024 & $\sim$ 8.29 \tabularnewline
\hline
512$\times$1024 (w/ LoRA) & $\sim$ 9.13 \tabularnewline
\hline
\end{tabular}
\end{table}
In this paper, our propose AnyRefill is based on large T2I model FLUX (12B).
To investigate the relationship between input size and inference cost, we provide the inference speed for different input resolutions in same codebase, shown in Tab.~\ref{tab:infer_speed}. 
All tests are based on 50 RF sampling steps.
LeftRefill needs to stitch two images together, which would double the input size. But the inference time only increases 3.03s, from 5.26s to 8.29s, as shown in Tab.~\ref{tab:infer_speed}. Meanwhile, incorporating task-specific LoRA into FLUX.Fill increases the model’s inference time by only 0.84 seconds, from 8.29s to 9.13s, which remains within an controllable range.
Therefore, we think the proposed LPG fomulation's inference cost is still acceptable in most real-world applications.

\subsection{Limitation}
\label{sec:limitation}
Although the proposed AnyRefill with LPG pattern enjoys good performance and reference alignment in various vision tasks, investigating the efficiency of multi-view generation for AnyRefill, which is based on large-scale T2I models like FLUX, can be regarded as interesting future work of LPG paradigm.
Moreover, The relationship between the amount of training data and model performance under the LPG paradigm is also a highly valuable direction for exploration.